\documentclass{article}

\usepackage[nonatbib,final]{neurips_2021}

\usepackage[square,sort,comma,numbers]{natbib}
\usepackage[utf8]{inputenc} 
\usepackage[T1]{fontenc}    
\usepackage{hyperref}       
\usepackage{url}            
\usepackage{booktabs}       
\usepackage{amsfonts}       
\usepackage{nicefrac}       
\usepackage{microtype}      
\usepackage{xcolor}         
\usepackage{graphicx}

\usepackage{amsmath}
\usepackage{bm}

\def\eqref#1{equation~\ref{#1}}

\def\1{\bm{1}}

\DeclareMathAlphabet{\mathsfit}{\encodingdefault}{\sfdefault}{m}{sl}
\SetMathAlphabet{\mathsfit}{bold}{\encodingdefault}{\sfdefault}{bx}{n}

\newcommand{\E}{\mathbb{E}}

\usepackage{url}
\usepackage{multirow}
\usepackage{optidef}
\usepackage{amsthm,amssymb}
\usepackage{subfig}
\usepackage{wrapfig}
\usepackage{pifont}

\usepackage[font=small,
            labelfont=it]{caption}
\newcommand{\ie}{\textit{i.e.}}
\newcommand{\eg}{\textit{e.g.}}

\newcommand{\best}[1]{
\textbf{#1}
}

\usepackage{lipsum}
\usepackage[ruled]{algorithm2e}
\usepackage{algorithmic}
\usepackage{colortbl}
\definecolor{na}{gray}{0.9}

\title{Gradient-based Editing of Memory Examples for Online Task-free Continual Learning}

\author{%
  \textbf{Xisen Jin} \quad \textbf{Arka Sadhu} \quad \textbf{Junyi Du} \quad \textbf{Xiang Ren} \\
  University of Southern California\\
  \texttt{\{xisenjin, asadhu, junyidu, xiangren@usc.edu\}} \\
}

\begin{document}

\maketitle

\begin{abstract}
We explore task-free continual learning (CL), in which a model is trained to avoid catastrophic forgetting in the absence of explicit task boundaries or identities.
Among many efforts on task-free CL, a notable family of approaches are memory-based that store and replay a subset of training examples.
However, the utility of stored seen examples may diminish over time since CL models are continually updated.
Here, we propose Gradient based Memory EDiting (GMED), a framework for editing stored examples in continuous input space via gradient updates, in order to create more ``challenging'' examples for replay. 
GMED-edited examples remain similar to their unedited forms, but can yield increased loss in the upcoming model updates, thereby making the future replays more effective in overcoming catastrophic forgetting. 
By construction, GMED can be seamlessly applied in conjunction with other memory-based CL algorithms to bring further improvement.
Experiments validate the effectiveness of GMED, and our best method significantly outperforms baselines and previous state-of-the-art on five out of six datasets\footnote{Code can be found at \url{https://github.com/INK-USC/GMED}.}. 

\end{abstract}
\vspace{-0.1cm}
\section{Introduction}
\label{sec:intro}

Learning from a continuous stream of data -- referred to as \textit{continual learning (CL)} or \textit{lifelong learning} -- has recently seen a surge in interest, and many works have proposed ways to mitigate CL models' catastrophic forgetting of previously learned knowledge~\citep{Robins1995CatastrophicFR, Lange2019ACL,Parisi2019ContinualLL}.
Here, we study online \textit{task-free} CL \citep{Aljundi2018TaskFreeCL}, where task identifiers and boundaries are absent from the data stream.
This setting reflects many real-world data streams ~\citep{Liu2020LearningOT,Caccia2020OnlineFA} and offers a challenging testbed for online CL research.


Memory-based methods, a prominent class of approaches used for task-free continual learning, 
store a small number of training examples (from the data stream) in a memory and replay them at the later training iterations~\citep{Robins1995CatastrophicFR, Rolnick2019ExperienceRF}. 
Existing methods operate over the original examples in the data-stream and focus on identifying samples to populate the memory \citep{Aljundi2019GradientBS, Chrysakis2020OnlineCL} and finding samples in the memory to be replayed ~\citep{Aljundi2019OnlineCL}.
However, for continually updating models, using stored-seen examples in their original form, may lead to diminishing utility over time --- i.e., model may gradually memorize the stored examples after runs of replay, as the memory refreshes slowly.
An alternate approach is to use generative models to create samples that suffers more from forgetting such as in GEN-MIR in~\citep{Aljundi2019OnlineCL}. 
In practice, training the generator network with limited data is challenging and leads to low-quality generated examples. Further, in the online learning setup, the generative model itself suffers from forgetting. 
As such, generative models perform worse than their memory counter-parts.

In this paper, we present a novel memory-based CL framework, Gradient based Memory EDiting (\textbf{GMED}), which looks to directly ``edit'' (via a small gradient update) examples stored in the replay memory.
These edited examples are stored (replacing their unedited counterparts), replayed, and further edited, thereby making the future replays more effective in overcoming catastrophic forgetting. 
Since no explicit generative model is involved, GMED approach retains the advantages of memory-based methods and is straightforward to train only inducing a small computation overhead.

The main consideration in allowing ``editing'' via a gradient update is the choice of the optimization objective.
In light of recent work on designing alternative replay strategies~\citep{Aljundi2019OnlineCL, toneva2018an, Chaudhry2020UsingHT}, we hypothesize that ``interfering'' examples (\ie, past examples that suffer from increased loss) should be prioritized for replay.
For a particular stored example, GMED finds a small update over the example (``edit'') such that the resulting edited example yields the most increase in loss when replayed.
GMED additionally penalizes the loss increase in the edited example to enforce the proximity of the edited example to the original sample, so that the edited examples stay in-distribution.
As a result, replaying these edited examples is more effective in overcoming catastrophic forgetting.
Since GMED focuses only on editing the stored examples, by construction, GMED is modular, \ie, it can be seamlessly integrated with other state-of-the-art memory-based replay methods~\citep{LopezPaz2017GradientEM, Aljundi2019OnlineCL, Buzzega2020RethinkingER}.

We demonstrate the effectiveness of GMED with a comprehensive set of experiments over six benchmark datasets.
In general, combining GMED with existing memory-based approaches results in consistent and statistically significant improvements with our single best method establishing a new state-of-art performance on five datasets.
Our ablative investigations reveal that the gains realized by GMED are significantly larger than those obtained from regularization effects in random perturbation, and can be accumulated upon data augmentation to further improve performance.

To summarize, our contributions are two-fold: 
(i) we introduce GMED, a modular framework for task-free online continual learning, to edit stored examples and make them more effective in alleviating catastrophic forgetting
(ii) we perform intensive set of experiments to test the performance of GMED under various datasets, parameter setups (\eg, memory size) and competiting baseline objectives.


\vspace{-0.1cm}
\section{Related Work}
\label{sec:related}
\vspace{-0.2cm}



\textbf{Continual Learning} studies the problem of learning from a data stream with changing data distributions over time~\citep{Lesort2020ContinualLF, Lange2019ACL}.
A major bottleneck towards this goal is the phenomenon of catastrophic forgetting \citep{Robins1995CatastrophicFR} where the model ``forgets'' knowledge learned from past examples when exposed to new ones.
To mitigate this effect, a wide variety of approaches have been investigated such as 
adding regularization 
~\citep{kirkpatrick2017overcoming, zenke2017continual, v.2018variational, Adel2020Continual},
separating parameters for previous and new data~\citep{Rusu2016ProgressiveNN, Serr2018OvercomingCF,Li2019LearnTG},
replaying examples from memory or a generative model
~\citep{Robins1995CatastrophicFR, Shin2017ContinualLW, LopezPaz2017GradientEM},
meta-learning 
~\citep{Javed2019MetaLearningRF}.
In this work, we build on memory-based approaches which have been more successful in the online task-free continual learning setting that we study. 

\noindent \textbf{Online Task-free Continual Learning}~\citep{Aljundi2018TaskFreeCL} 
is a specific formulation of the continual learning where the task boundaries and identities are not available to the model. 
Due to its broader applicability to real-world data-streams, a number of algorithms have been adapted to the task-free setup \citep{Zeno2018TaskAC,Aljundi2019OnlineCL,harrison2019continuousmw,He2019TaskAC,Lee2020A}.
In particular, memory-based CL algorithms which store a subset of examples and later replay them during training, have seen elevated success in the task-free setting.
Improvements in this space have focused on: \textit{storing diverse examples} as in Gradient-based Sample Selection (GSS)~\citep{Aljundi2019GradientBS}, and \textit{replaying examples with larger estimated ``interference''} as in Maximally Interfered Retrieval (MIR) with experience replay~\citep{Aljundi2019OnlineCL}.
In contrast 
, GMED is used in conjunction with memory-based approaches and explicitly searches for an edited example  which is optimized to be more ``effective'' for replay.


\textbf{Replay Example Construction.} Moving away from replaying real examples, a line of works on \textit{deep generative replay}~\citep{Shin2017ContinualLW, Hu2019OvercomingCF, Rostami2019ComplementaryLF} generates synthetic examples to replay with a generative model trained online. GEN-MIR~\citep{Aljundi2019OnlineCL} is further trained to generate examples that would suffer more from interference for replay. 
However, training a generative network is challenging, even more so in the online continual learning setup where the streaming examples are encountered only once leading to poorly generated examples.
Moreover, the forgetting of generative networks themselves cannot be perfectly mitigated.
As a result, these methods generally perform worse than their memory-based counter-parts.
Instead of generating a new example via a generator network, GMED uses gradient updates to directly edit the stored example thereby retaining advantages of memory-based techniques while creating new samples.

Novel examples can also be constructed via \textit{data augmentation} (\eg{} random crop, horizontal flip) to help reduce over-fitting over the small replay memory~\citep{Buzzega2020RethinkingER}. Unlike GMED, these data-augmentations are usually pre-defined and cannot adapt to the learning pattern of the model.
Constructing edited examples has also been used for adversarial robustness~\citep{Song2018ConstructingUA, Papernot2017PracticalBA}.
The key difference lies in the optimization objective: while adversarial construction focuses on finding mis-classified examples~\citep{Goodfellow2015ExplainingAH, madry2018towards}, GMED aims at constructing ``interfered'' examples that would suffer from loss increase in future training steps.

\vspace{-0.1cm}
\section{Method}
\label{sec:method}
\vspace{-0.2cm}
We introduce the formulation of online task-free continual learning (Sec. \ref{sec:background}) 
and then present our gradient-based memory editing method GMED (Sec. \ref{ssec:gmed_main}) and detail how GMED can be integrated with experience replay (Sec ~\ref{ssec:gmed_with_er}) and other memory-based CL algorithms (Sec. ~\ref{ssec:gmed_mir}).

\begin{figure*}
\vspace{-0.2cm}
    \centering
    \includegraphics[width=\linewidth]{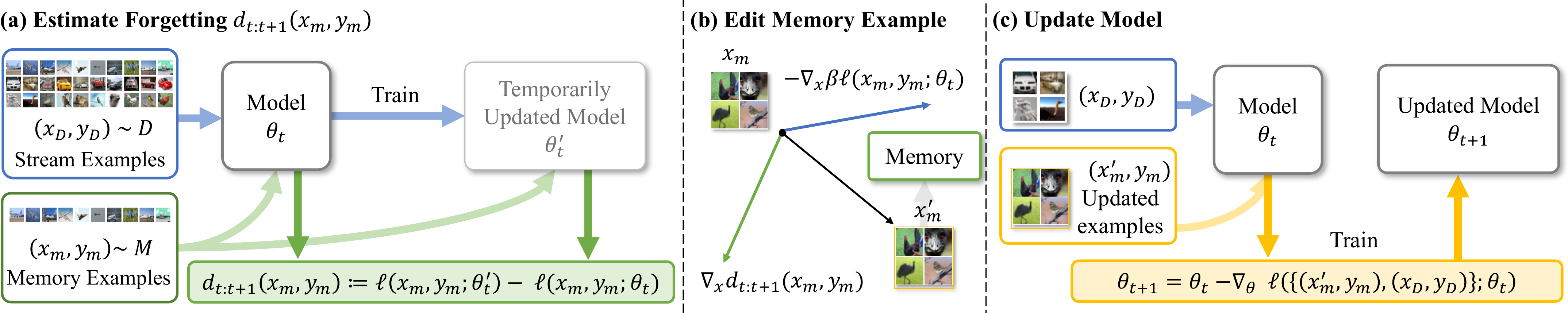}
    \vspace{-0.2cm}
    \caption{\small \textbf{A schematic of GMED framework}. 
    (a) Given an example from the data-stream $(x_D, y_D)$  at time $t$, the model randomly draws an example from the memory $(x_m, y_m)$ and estimates ``interference'' after one-step roll-out (Eq. \ref{eq:estimate_forgetting}).
    (b) The example drawn from the memory is then updated using our proposed editing objective (Eq. \ref{eq:editing}) via gradient ascent, resulting in $(\hat{x}_m,y_m)$, and written back into the memory.
    (c) Finally, the model is updated using the edited example $(\hat{x}_m,y_m)$ and the example from the data-stream $(x_D, y_D)$ (Eq. ~\ref{eqn:theta_update}).
    }
    \label{fig:overview} 
    \vspace{-0.4cm}
\end{figure*}

\vspace{-0.2cm}
\subsection{Problem Formulation}
\label{sec:background}
\vspace{-0.2cm}



In continual learning (CL), we consider a (potentially infinite) stream $D$ of labeled examples $(x,y)$, having a non-stationary data distribution -- \ie, the data distribution $p(x,y)$ evolves over time.
Let $(x_t, y_t)$ denote the labeled example (or a mini-batch of examples) received by the model at time-step $t$ in the data-stream $D$.
We assume, for simplicity, that $(x_t, y_t)$ is generated by first sampling a \textit{latent} ``task'' $z \sim p(z; t)$, followed by sampling a data example from a conditional data distribution $p(x, y|z)$, -- \textit{i.e.}, $(x_t, y_t) \sim p(x, y|z)$. 
Here $p(z; t)$ is non-i.i.d and time-dependent. 
In \textit{task-free} CL setting, the latent task $z$ is \textit{not} revealed to the model.
Finally, we emphasize that  
while models in task-aware setup proceed to the next task only after convergence on the current task~\citep{kirkpatrick2017overcoming, zenke2017continual}, this work focuses on \textbf{\textit{online} task-free CL} setup~ \citep{chaudhry2018efficient, Chrysakis2020OnlineCL}. It simulates a practical setting where models must perform online update on every incoming example, without accumulating examples within a task.

Following above definitions, our goal is to learn a classification model $f(x; \theta)$ on the data stream $D$ that can preserve its performance over all the tasks in $D$.
At time step $t$ during the training process, the model is updated to minimize a predefined empirical loss $\ell(x_t, y_t; \theta)$ on the newly received example $(x_t, y_t)$, without increasing the loss on the previously visited examples (before $t$). Specifically, let $p_c(x,y;T)$ denotes the distribution of the examples visited until the time step $T$. We look to minimize the expected loss $\E_{p_c(x,y;T)} \ell(x,y; \theta)$ -- \textit{i.e.}, retaining performance on tasks encountered before $T$.

\subsection{Gradient based Memory Editing (GMED)}
\label{ssec:gmed_main}
\vspace{-0.2cm}



In online task-free continual learning, examples visited earlier cannot be accessed (revisited) and thus computing the loss over all the visited examples (in $D$) is not possible. 
To deal with this challenge, memory-based CL algorithms store and (continuously) maintain a set of visited examples in a fixed-size memory and use them for replay or regularization in the future training steps ~\citep{Robins1995CatastrophicFR,LopezPaz2017GradientEM}. 
For instance, in Experience Replay (ER) ~\citep{Robins1995CatastrophicFR} the examples are randomly sampled from the memory for replay; whereas recent works explore more sophisticated replay strategies, such as Maximally Interfered Retrieval with experience replay (ER-MIR)~\citep{Aljundi2019OnlineCL} where
memory examples which interfere the most with the newly visited example are selected for replay. 
However, these algorithms only train over samples drawn from a small replay memory in their original form. As such, the utility of the stored examples could diminish over time as the model could potentially memorize these examples if the memory examples are refreshed in a slow pace -- which is found often the case~\citep{Buzzega2020RethinkingER}.

We address the above limitation in our proposed approach Gradient based Memory Editing (GMED) by allowing examples stored in the memory to be edited in the continuous input space, illustrated in Figure~\ref{fig:overview}.
The editing step is guided by an optimization objective instead of being pre-defined and involves drawing examples from the memory, editing the drawn examples, replaying the edited examples and at the same time writing the edited examples back to the memory.
We now state our optimization objectives of example editing followed by algorithmic details of GMED.

\begin{wrapfigure}[27]{R}{0.5\textwidth}
\begin{algorithm}[H]
\begin{small}
\caption{\small Gradient Memory EDiting with ER (ER+GMED)}
\begin{algorithmic}[1]
   \STATE {\bfseries Input:} learning rate $\tau$, edit stride $\alpha$, regularization strength $\beta$, decay rate $\gamma$, model parameters $\theta$
   \STATE {\bfseries Receives}: stream example $(x_D, y_D)$
   \STATE {\bfseries Initialize}: replay memory $M$
      
    \smallskip
   \FOR{$t=1$ {\bfseries to} $T$}
      \smallskip
   \STATE \texttt{\small //when $\gamma=1$, $k$ is not required}
   \STATE $(x_m,y_m) \sim M$;~~$k \leftarrow  \textrm{replayed\_time} (x_m,y_m)$
   \STATE $\ell_\textrm{before}\leftarrow \textrm{loss}(x_m,y_m,\theta_t)$;
  \STATE $\ell_\textrm{stream}\leftarrow \textrm{loss}(x_D, y_D,\theta_t)$

   \medskip
   \STATE \texttt{\small//update model with stream examples}
      \smallskip
   \STATE $\theta_t^\prime \leftarrow  \textrm{SGD}(\ell_\textrm{stream}, \theta_t, \tau)$ 

    \medskip
   \STATE \texttt{\small//evaluate forgetting of memory examples}
   \STATE $\ell_\textrm{after}\leftarrow \textrm{loss}(x_m,y_m,\theta_t^\prime)$
   \STATE $d\leftarrow \ell_\textrm{after} - \ell_\textrm{before}$
   
   \medskip
   \STATE \texttt{\small//edit memory examples}
      \smallskip
   \STATE $x_m^\prime \leftarrow x_m + \gamma^k \alpha \nabla_x (d - \beta \ell_\textrm{before})$

   \STATE $\ell = \textrm{loss}(\{(x_m^\prime,y_m), (x_D, y_D)\}, \theta_t)$
   \smallskip
   \STATE $\theta_{t+1} \leftarrow \textrm{SGD}(\ell, \theta_t, \tau)$
      \smallskip
   \STATE replace $(x_m, y_m)$ with $(x_m^\prime, y_m)$ in $M$
      \smallskip
   \STATE $\textrm{reservoir\_update}(x_D, y_D, M) $
      \smallskip
   \ENDFOR
\end{algorithmic}
\label{algo:gmed}
\end{small}
\end{algorithm}
\end{wrapfigure}

\textbf{Editing Objective.} 
Clearly, the most crucial step involved in GMED is identifying ``\textit{how}'' should the stored examples in the memory be edited to reduce catastrophic forgetting of early examples. If $(x_m,y_m)$ denotes an example drawn from the memory $M$, the goal is to design a suitable editing function $\phi$ to generate the edited  example $\hat{x}_m$ where $\hat{x}_m {=} \phi(x_m)$. 
Such ``editing'' process can also be found in white-box adversarial example construction for robustness literature~\citep{Goodfellow2015ExplainingAH} where the task is to find an ``adversarial'' example $(x^\prime, y)$ that is close to an original example $(x,y)$ but is mis-classified by the model. Typically, adversarial example construction utilizes the gradients obtained from the classifier and edits the examples to move towards a different target class.
While, in the context of continual learning, the editing objective should be different. We employ a similar hypothesis as previous works~\citep{Aljundi2019OnlineCL,toneva2018an, Chaudhry2020UsingHT} that examples that are likely forgotten by models should be prioritized for replay. Accordingly, we edit examples so that they are more likely to be forgotten in future updates. With this objective of editing in place, we detail the process of incorporating GMED with Experience Replay (ER). 
\subsection{The GMED Algorithm with ER}
\label{ssec:gmed_with_er}
Algorithm~\ref{algo:gmed} summarizes the process and Figure~\ref{fig:overview} provides a schematic of the steps involved in GMED. At time step $t$, the model receives a mini-batch of the stream examples $(x_D, y_D)$ from the training stream $D$, and randomly draws a same number of memory examples $(x_m^k, y_m)$ from the memory $M$, which we assume has already been drawn for replay for $k$ times.
We first compute the ``interference'' (\textit{i.e.,} loss increase) on the memory example $(x_m^k, y_m)$ when the model performs one gradient update on parameters with the stream example $(x_D, y_D)$.
\begin{gather}
\label{eq:gmed_update}
 \theta^\prime_t = \theta_t - \nabla_{\theta}\ell(x_D, y_D; \theta_t); \\
\begin{split}
 \label{eq:estimate_forgetting}
    d_{t:t+1}(x_m^k, y_m) &= \ell(x_m^k, y_m; \theta^\prime_t) -     \ell(x_m^k, y_m; \theta_t),
\end{split}
\end{gather}
where $\theta_t$ and $\theta_t^\prime$ are model parameters before and after the gradient update respectively. Then, we perform a step of gradient update on $(x_m^k, y_m^k)$ by maximizing its ``loss increase'' in the next one step of training, while using a regularization term $\ell(x_m, y_m; \theta_t)$ to penalize the loss increase evaluated with the current model checkpoint.
The iterative update is written as,
\begin{equation}
\label{eq:editing}
    x_m^{k+1}  \leftarrow x_m^{k} + \gamma^k \alpha \nabla_{x} [d_{t:t+1}(x_m^{k}, y_m) - \beta \ell (x_m^{k},y_m;\theta_t)],
\end{equation}
where the hyper-parameter $\alpha$ controls the overall stride of the edit and is tuned with first three tasks together with the regularization strength $\beta$. $\gamma$ is a decay factor of edit performed on the model. A decay factor $\gamma$ less than 1.0 could effectively prevent $x_m^k$ from drastically deviating from their original state, while $\gamma=1.0$ indicates no decay. We note that we cannot perform constrained edits that strictly adhere to a distance budget w.r.t original examples, as it requires storing original examples which introduces extra memory overhead.

Following Eq.~\ref{eq:editing}, we perform a gradient update on $x$ to increase its ``interference''.
The algorithm then discards $\theta_t^\prime$, and updates model parameters $\theta_t$ using the edited memory example $(x^{k+1}_m, y_m)$ and the stream example $(x_D, y_D)$, in a similar way to ER. 
\begin{equation}
    \label{eqn:theta_update}
    \theta_{t+1} = \theta_t - \nabla_\theta\ell(\{(x^{k+1}_m, y_m), (x_D, y_D)\};\theta_t).
\end{equation}
We replace the original examples in the memory with the edited example. 
In this way, we continuously edit examples stored in the memory alongside training.

\subsection{Applying GMED with Data Augmentation, MIR and GEM} 
\label{ssec:gmed_mir}
Since the process to edit the original examples in GMED is modular, we can integrate GMED with a range of existing memory-based CL algorithms.
In addition to ER, we also explore ER with data augmentation (ER$_{aug}$)~\citep{Buzzega2020RethinkingER}, ER-MIR  ~\citep{Aljundi2019OnlineCL} and GEM ~\citep{LopezPaz2017GradientEM} in our experiments.

ER$_{aug}$ applies standard data augmentations (\eg, random cropping, horizontal flipping, denoted as $\mathcal{T}$) to examples $(x_m, y_m)$ drawn from the memory which are replayed at each time step. 
In ER$_{aug}+$GMED, we edit original example $x_m$ and replay both the augmented example $\mathcal{T}(x_m)$ and the edited example $\hat{x}_m$. To keep the number of replayed examples the same, for ER$_{aug}$ method, we also replay both the edited example $\mathcal{T}(x_m)$ and the original example $x_m$. Finally, we write the edited example $\hat{x}_m$ to the memory.   

For integration with ER-MIR (denoted henceforth as MIR for brevity), recall that MIR retrieves and then replays the most ``interfering'' examples
in the memory at each time-step. 
Thus, making GMED edits on the MIR-retrieved examples may induce a loop of further increasing ``interference'' of the examples that are already the most ``interfering'' ones.
Instead, in our MIR$+$GMED implementation, we edit a  mini-batch of examples randomly sampled from the memory
--- a process that is \textit{independent} of the MIR replay operation\footnote{This also ensures the integrated approach will replay the same number of examples as the baselines, yielding a fair comparison.}.
This random sampling may help prevent GMED editing from intensifying potential biases created from the MIR retrieval process (\textit{e.g.}, retrieved examples are edited and thus become more interfered).
Similarly, For GEM$+$GMED, we also apply GMED to edit a mini-batch of randomly sampled examples.

Details of the integrated algorithms (\textit{i.e.}, ER$_{aug}+$GMED, MIR$+$GMED, and GEM$+$GMED) can be found in Algorithms~\ref{algo:gmed_er_aug},~\ref{algo:gmed_mir} and~\ref{algo:gmed_gem} in Appendix~\ref{apdx:algo_details_mir_gem}.
We leave more sophisticated integration of GMED to existing CL algorithms \eg{} by optimizing the retrieval and editing operations jointly to future work.

\vspace{-0.2cm}
\section{Experiments}
\label{sec:exps}
\vspace{-0.2cm}

Our experiments address the following research questions: 
(i) what are the gains obtained by integrating GMED with existing memory-based CL algorithms and how these gains compare across datasets, methods and memory sizes?
(ii) how useful are GMED edits in alleviating catastrophic forgetting, and what part of it can be attributed to the design of the editing objective function?
(iii) what role do the various components and parameters in the GMED play and how they affect the performance.
In the rest of this section, we first briefly describe the datasets (Sec.~\ref{ssec:datasets}) and the compared baselines (Sec.~\ref{ssec:compared_methods}).
We then detail our main results comparing across memory-based algorithms (Sec.~\ref{ssec:results}), validate the effectiveness of GMED-edits and the editing objective in mitigating catastrophic forgetting  (Sec. ~\ref{ssec:gmed_abl_optimal}), followed by ablative study over its components (Sec. ~\ref{ssec:gmed_abl_params}). 









\vspace{-0.2cm}
\subsection{Datasets}
\label{ssec:datasets}
\vspace{-0.1cm}
We use six public CL datasets in our experiments. \textbf{Split / Permuted / Rotated MNIST} are constructed from the MNIST~\citep{lecun1998mnist} dataset which contains images of handwritten digits.
Split MNIST~\citep{goodfellow2013empirical}
creates 5 disjoint subsets based on the class labels and considers each subset as a separate task.
The goal then is to classify over all 10 digits when the training ends. 
Permuted MNIST ~\citep{goodfellow2013empirical} 
consists of 10 tasks, where for a particular task a random permutation in the pixel space is chosen and applied to all images within that task.
The model then has to classify over the 10 digits without knowing which random permutation was applied. 
Rotated MNIST
~\citep{LopezPaz2017GradientEM} rotates every sample in MNIST by a fixed angle between 0 to 180.
Similar to the previous datasets,
the goal is to classify over 10 digits without any knowledge of the angle of rotation. 
For all MNIST experiments, each task consists of 1,000 training examples following ~\citep{Aljundi2019OnlineCL}.
We also employ \textbf{Split CIFAR-10 and Split CIFAR-100}, which comprise of 5 and 20 disjoint subsets respectively based on their class labels.
The model then classifies over the space of all class labels.
Similarly, \textbf{Split mini-ImageNet} ~\citep{Aljundi2019OnlineCL} splits the mini-ImageNet~\citep{deng2009imagenet,Vinyals2016MatchingNF} dataset into 20 disjoint subsets based on their labels. 
The models classify over all 100 classes.

Following the taxonomy of~\citep{Ven2019ThreeSF}, the Split MNIST, Split CIFAR-10, Split CIFAR-100, and Split mini-ImageNet experiments are categorized under class-incremental setup, while Permuted and Rotated MNIST experiments belong to domain-incremental setup. We note that our results are not comparable to works that employ a different setup over the same dataset (\eg, results on domain-incremental Split CIFAR-100 are not comparable).

\vspace{-0.2cm}
\subsection{Compared Methods}
\label{ssec:compared_methods}
\vspace{-0.1cm}
We compare against several task-free memory based continual learning methods namely, 
Experience Replay (ER) ~\citep{Robins1995CatastrophicFR}, 
Averaged Gradient Episodic Memory (AGEM)~\citep{chaudhry2018efficient},
Gradient based Sample Selection (GSS)~\citep{Aljundi2019GradientBS}, and Maximally Interfering Retrieval (MIR)~\citep{Aljundi2019OnlineCL}. 
We omit the generative replay method GEN-MIR proposed together in~\citep{Aljundi2019OnlineCL} as it underperforms their memory-based counterparts even on simple datasets such as Split MNIST.

We also compare with data augmentation ~\citep{Buzzega2020RethinkingER} such as random rotations, scaling, and horizontal flipping applied to memory examples drawn for replay in ER, noted as ER$_{aug}$ (except for MNIST datasets).
We also include regularization-based, model expansion-based and task-aware approaches, namely Bayesian Graident Descent (BGD)~\citep{Zeno2018TaskAC}, Neural Dirichlet Process Mixture Model (CN-DPM)~\citep{Lee2020A}, Progressive Networks (Prog.NN)~\citep{Rusu2016ProgressiveNN}, Compositional Lifelong Learning~\citep{mendez2021lifelong}, Graident Episodic Memory (GEM)~\citep{LopezPaz2017GradientEM} and Hindsight Anchor Learning (HAL)~\citep{Chaudhry2020UsingHT} respectively.

Finally,
we report three baseline models:
(i) Fine Tuning,
where no continual learning algorithms are used for online updates to model parameters,
(ii) iid Online, where we randomly shuffle the data stream, so that the model visits an i.i.d. stream of examples, and 
(iii) iid Offline, where multiple passes over the dataset is allowed. 
Appendix~\ref{sec:details_of_baselines} provides more 
details on compared methods and their implementation details. 
We build our proposed GMED approach upon four baselines, noted as ER+GMED, MIR+GMED, GEM+GMED, and ER$_{aug}$+GMED.



\begin{table*}[tb]
\vspace{-0.5cm}
\caption{\small \textbf{Mean and standard deviation of final accuracy (\%) for non-model-expansion-based approaches on $6$ datasets}. 
For Split mini-ImageNet and Split CIFAR-100 datasets, we set the memory size to 10,000 and 5,000 examples; we use 500 for other datasets. $^{*}$ and $^{**}$ over GMED methods indicate significant improvement over the counterparts without GMED with $p$-values less than $0.1$ and $0.05$ respectively in single-tailed paired t-tests. We report results in 20 runs for GEM, ER, MIR, and ER$_{aug}$ and their GMED-integrated versions, and 10 runs for others. $\dagger$ over ``previous SOTA'' results indicates that the best GMED method (\textbf{bolded} for each dataset) outperforms the previous SOTA with statistically significant improvement ($p<0.1$). 
}
\label{tab:main_tab}
\centering
\scalebox{0.70}{\begin{tabular}{@{}lcccccc@{}}
\toprule
\textbf{Methods / Datasets} & \textbf{\begin{tabular}[c]{@{}c@{}}Split\\ MNIST\end{tabular}} & \textbf{\begin{tabular}[c]{@{}c@{}}Permuted\\ MNIST\end{tabular}} & \textbf{\begin{tabular}[c]{@{}c@{}}Rotated\\ MNIST\end{tabular}} & \textbf{\begin{tabular}[c]{@{}c@{}}Split\\ CIFAR-10\end{tabular}} & \textbf{\begin{tabular}[c]{@{}c@{}}Split\\ CIFAR-100\end{tabular}} & \textbf{\begin{tabular}[c]{@{}c@{}}Split\\ mini-ImageNet\end{tabular}} \\ \midrule
Fine tuning & 18.80 $\pm$ 0.6  & 66.34 $\pm$ 2.6 & 41.24 $\pm$ 1.5 & 18.49 $\pm$ 0.2 & 3.06 $\pm$ 0.2 & 2.84 $\pm$ 0.4  \\
AGEM~\citep{chaudhry2018efficient} & 29.02 $\pm$ 5.3 & 72.17 $\pm$ 1.5 & 50.77 $\pm$ 1.9 &   18.49 $\pm$ 0.6 & 2.40 $\pm$ 0.2 & 2.92 $\pm$ 0.3  \\
GSS-Greedy~\citep{Aljundi2019GradientBS} &  84.16 $\pm$ 2.6 & 77.43 $\pm$ 1.4 & 73.66 $\pm$ 1.1  & 28.02 $\pm$ 1.3 & 19.53 $\pm$ 1.3 & 16.19 $\pm$ 0.7 \\
BGD~\citep{Zeno2018TaskAC} &  13.54 $\pm$ 5.1 & 19.38 $\pm$ 3.0 & 77.94 $\pm$ 0.9 & 18.23 $\pm$ 0.5 & 3.11 $\pm$ 0.2 & 24.71 $\pm$ 0.8 \\
\midrule
ER~\citep{Robins1995CatastrophicFR} & 81.07 $\pm$ 2.5                                                & 78.65 $\pm$ 0.7                                                   & 76.71 $\pm$ 1.6                                                  & 33.30 $\pm$ 3.9                               &     20.11 $\pm$ 1.2                    & 25.92 $\pm$ 1.2                                                        \\
\rowcolor{na} ER + GMED    &  82.67$^{**}$ $\pm$ 1.9                                           & 78.86 $\pm$ 0.7                                                  & 77.09$^{*}$ $\pm$ 1.3                                              & 34.84$^{**}$ $\pm$ 2.2                              &      20.93$^{*}$ $\pm$ 1.6          & 27.27$^{**}$ $\pm$ 1.8                                                    \\
\midrule
MIR~\citep{Aljundi2019OnlineCL}   & 85.72 $\pm$ 1.2                                                & 79.13 $\pm$ 0.7                                                   & 77.50 $\pm$ 1.6                                                  & 34.42 $\pm$ 2.4                          & 20.02 $\pm$ 1.7                      & 25.21 $\pm$ 2.2 \\
\rowcolor{na} MIR + GMED  &   \best{86.52$^{**}$ $\pm$ 1.4}                                          & \best{79.25 $\pm$ 0.8}                                               & 79.08$^{**}$ $\pm$ 0.8                                           & 36.17$^{*}$ $\pm$ 2.5       &                            \best{21.22$^{**}$ $\pm$ 1.0}            & 26.50$^{**}$ $\pm$ 1.3               \\
\midrule
ER$_{aug}$~\citep{Buzzega2020RethinkingER} &  80.14 $\pm$ 3.2                                                 &         78.11 $\pm$ 0.7                               & 80.04 $\pm$ 1.3                                      & 46.29 $\pm$ 2.7                       &         18.32 $\pm$ 1.9                  &  30.77 $\pm$ 2.2      
\\
\rowcolor{na} ER$_{aug}$ + GMED   &   82.21$^{**}$ $\pm$ 2.9                                                &            78.13 $\pm$ 0.6                         & \best{80.61$^{*}$ $\pm$ 1.2}                                                 &   \best{47.47$^*$ $\pm$ 3.2}                  &         19.60 $\pm$ 1.5                  & \best{31.81$^{*}$ $\pm$ 1.3}   \\
\midrule
Previous SOTA  & 85.72$^{\dagger}$ $\pm$ 1.2 &   79.23 $\pm$ 0.7  & 80.04$^{\dagger}$ $\pm$ 1.3 & 46.29$^{\dagger}$ $\pm$ 2.7 & 20.02$^{\dagger}$ $\pm$ 1.7  & 30.77$^{\dagger}$ $\pm$ 2.2   \\ 
iid online & 85.99 $\pm$ 0.3 &   73.58 $\pm$ 1.5  & 81.30 $\pm$ 1.3 &  62.23 $\pm$ 1.5 & 18.13 $\pm$ 0.8  &  17.53 $\pm$ 1.6   \\
iid offline (upper bound)  & 93.87 $\pm$ 0.5 &  87.40 $\pm$ 1.1  & 91.38 $\pm$ 0.7  & 76.36 $\pm$ 0.9 & 42.00 $\pm$ 0.9 & 37.46 $\pm$ 1.3 \\ 
\bottomrule
\end{tabular}
}
\vspace{-0.2cm}
\end{table*}





\noindent \textbf{Implementation Details}. 
We set the size of replay memory as $10K$ for split CIFAR-100 and split mini-ImageNet, and $500$ for all remaining datasets.
Following~\citep{chaudhry2018efficient}, we tune the hyper-parameters $\alpha$ (editing stride) and $\beta$ (regularization strength) with only the first three tasks. While $\gamma$ (decay rate of the editing stride) is a hyper-parameter that may flexibly control the deviation of edited examples from their original states, we find $\gamma{=} 1.0$ (\ie, no decay) leads to better performance in our experiments. 
Results under different $\gamma$ setups are provided in Appendix~\ref{apdx:hyp_setup}, and in the remaining sections we assume no decay is applied.
For model architectures, we mostly follow the setup of~\citep{Aljundi2019OnlineCL}:
for the three MNIST datasets, we use a MLP classifier with 2 hidden layers with 400 hidden units each. 
For Split CIFAR-10, Split CIFAR-100 and Split mini-ImageNet datasets, we use a ResNet-18 classifier with
three times less feature maps across all layers.
See Appendix~\ref{apdx:hyp_setup} for more details.


\vspace{-0.1cm}
\subsection{Performance Across Datasets}
\label{ssec:results}
\vspace{-0.2cm}

We summarize the results obtained by integrating GMED with different CL algorithms.
In Table~\ref{tab:main_tab}, we report the final accuracy and the standard deviation and make the following key observations.

\begin{wraptable}{R}{0.52\textwidth}
\vspace{-0.5cm}
\centering
\caption{\small \textbf{Comparison with model-expansion-based approaches} under the same memory overhead as CN-DPM. The overhead is the size of the replay memory plus the extra model components (e.g. a generator or modules to solve individual tasks), shown in the equivalent number of memory examples (\#. Mem). $\dagger$ indicates quoted numbers are taken from the respective papers. \\}
\vspace{-0.4cm}
\scalebox{0.67}{
\begin{tabular}{@{}lcccccc@{}}
\toprule
\textbf{Method}  & \multicolumn{2}{c}{\textbf{Split MNIST}} & \multicolumn{2}{c}{\textbf{Split CIFAR-10}} & \multicolumn{2}{c}{\textbf{Split CIFAR-100}} \\ \midrule
\textbf{}        & Acc.        & \#. Mem          & Acc.       & \#. Mem  & Acc. &  \#. Mem   \\ \midrule
\textbf{CN-DPM}$^\dagger$  & 93.23       & 2,581               & 45.21      & 6,024     & 20.10 & 21,295  \\
\textbf{Prog. NN} & 89.46 & 9,755  & 49.68 & 6,604 & 19.17 & 31,766 \\
\textbf{CompCL} & 91.27 & 9,755 & 45.62 & 6,604 & 20.51 & 31,766 \\ 
\textbf{ER}      & 92.67       & 2,581                & 62.96      & 6,024    & 21.79 &   21,295    \\
\textbf{ER+GMED} & \best{94.16}       & 2,581             & \best{63.28 }     & 6,024     & \best{22.12} &   21,295    \\
\bottomrule 
\end{tabular}
}
    \label{tab:comp_cndpm}
\vspace{-0.1cm}
\end{wraptable}

\noindent \textbf{Effectiveness of Memory Editing}.
As can be seen in Table~\ref{tab:main_tab}, GMED significantly improves performance on 5 datasets when built upon ER and MIR respectively. The improvement of MIR+GMED corroborates that the optimization in the continuous input spaces of GMED is complementary to sample selection over real examples as in MIR. We also notice significant improvement of ER$_{aug}$ over GMED on 5 datasets. This indicates that the benefits induced by GMED go beyond regularization effects used to mitigate over-fitting.

\begin{figure*}
\vspace{-0.3cm}
    \centering
    \subfloat[][Split MNIST]{\includegraphics[width=0.20\textwidth]{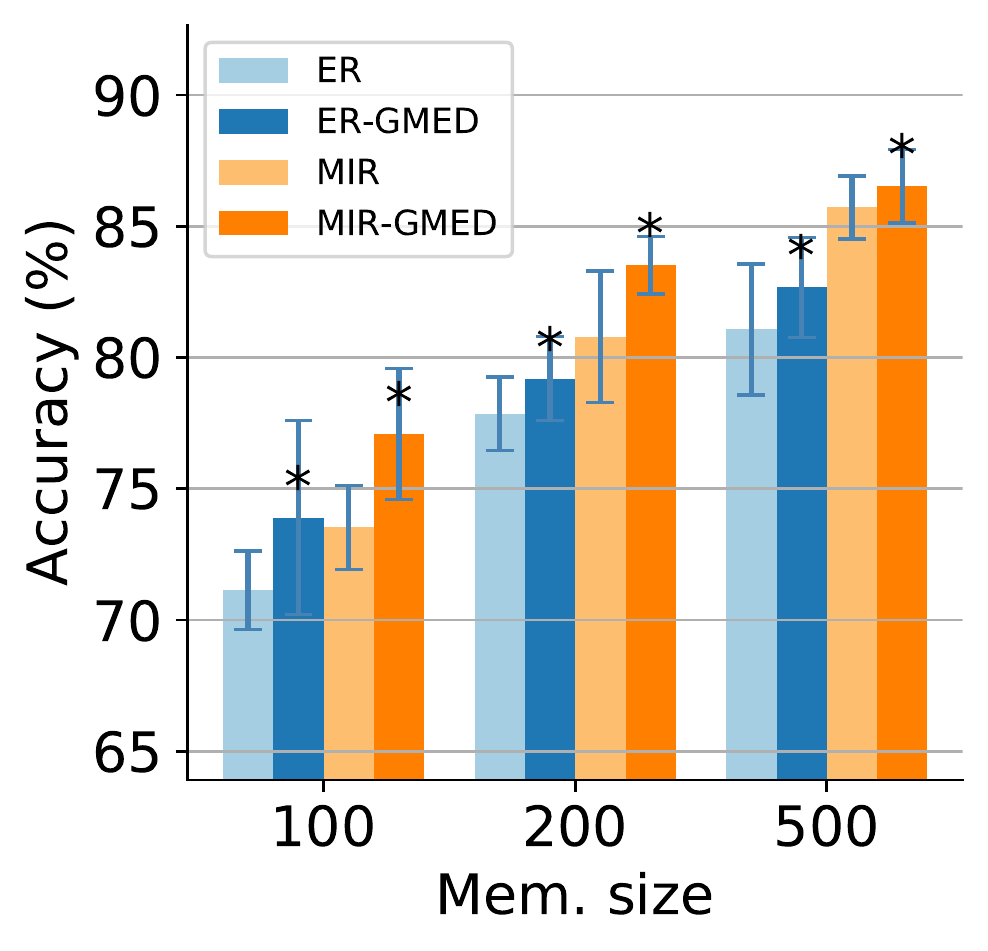}}
    \subfloat[][Rotated MNIST]{\includegraphics[width=0.20\textwidth]{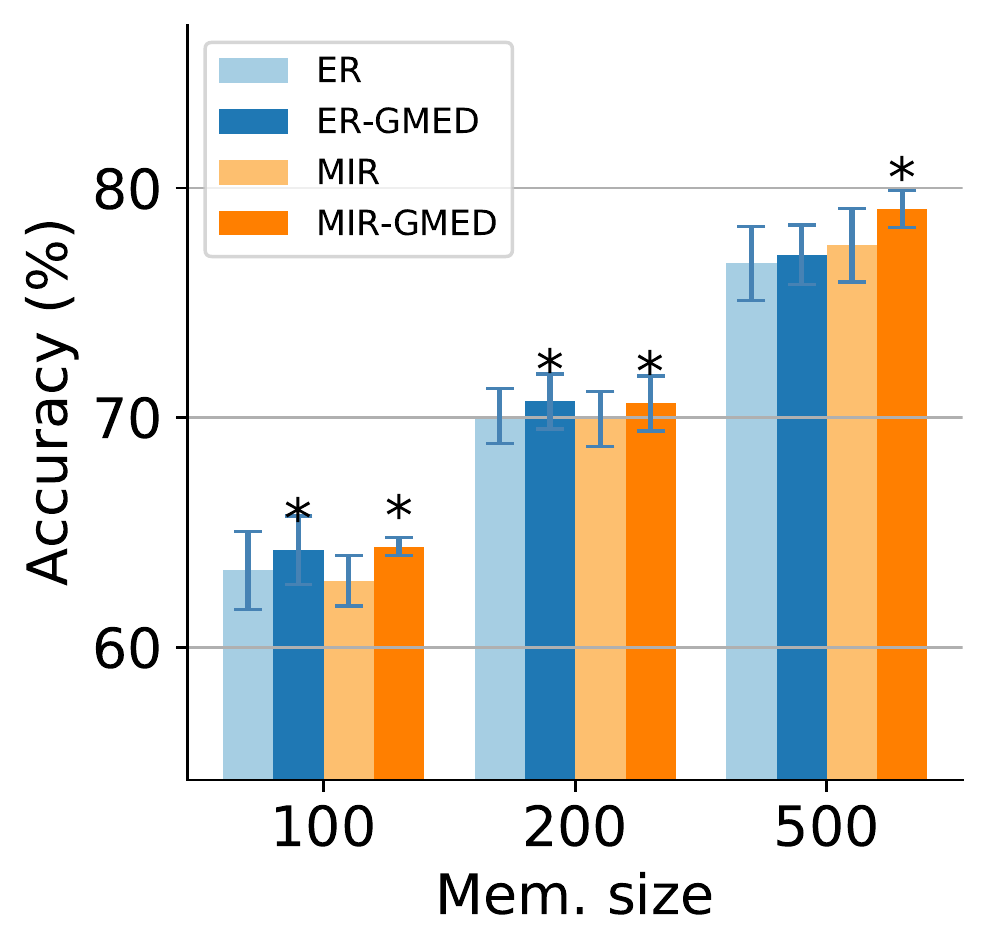}}
    \subfloat[][Split CIFAR-10]{\includegraphics[width=0.20\textwidth]{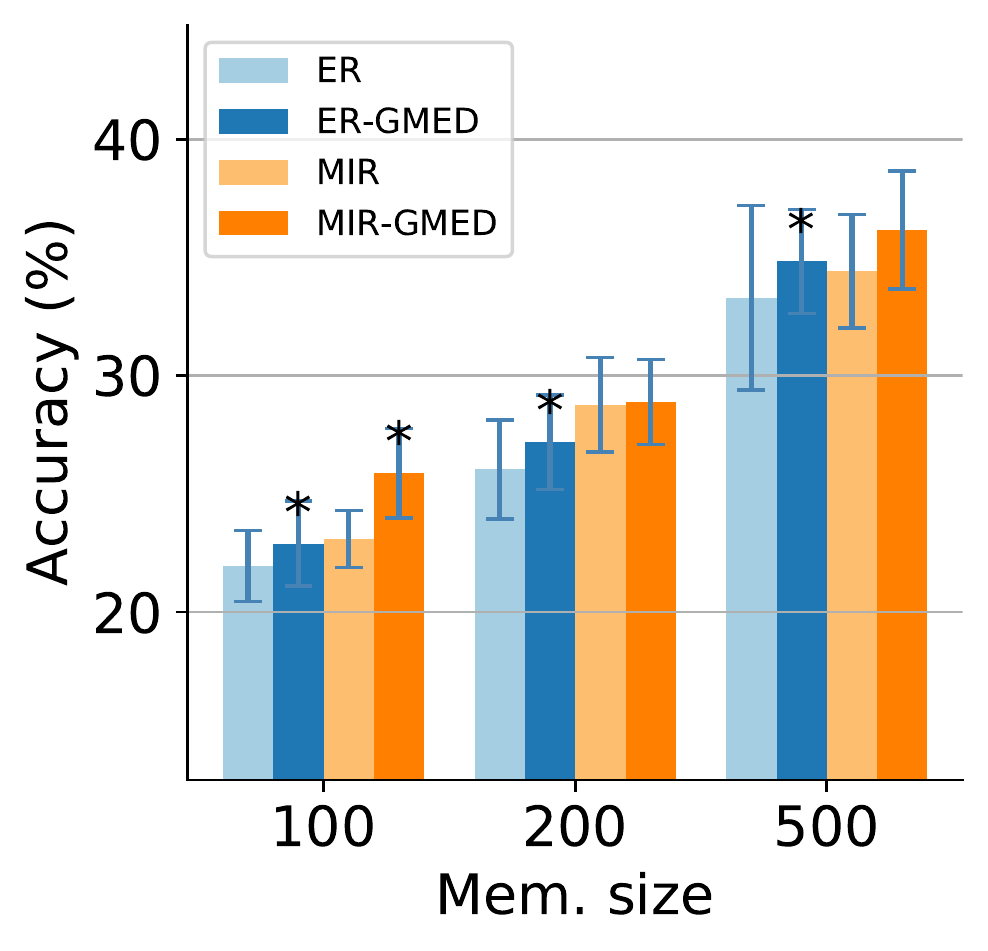}}
    \subfloat[][Split CIFAR-100]{\includegraphics[width=0.20\textwidth]{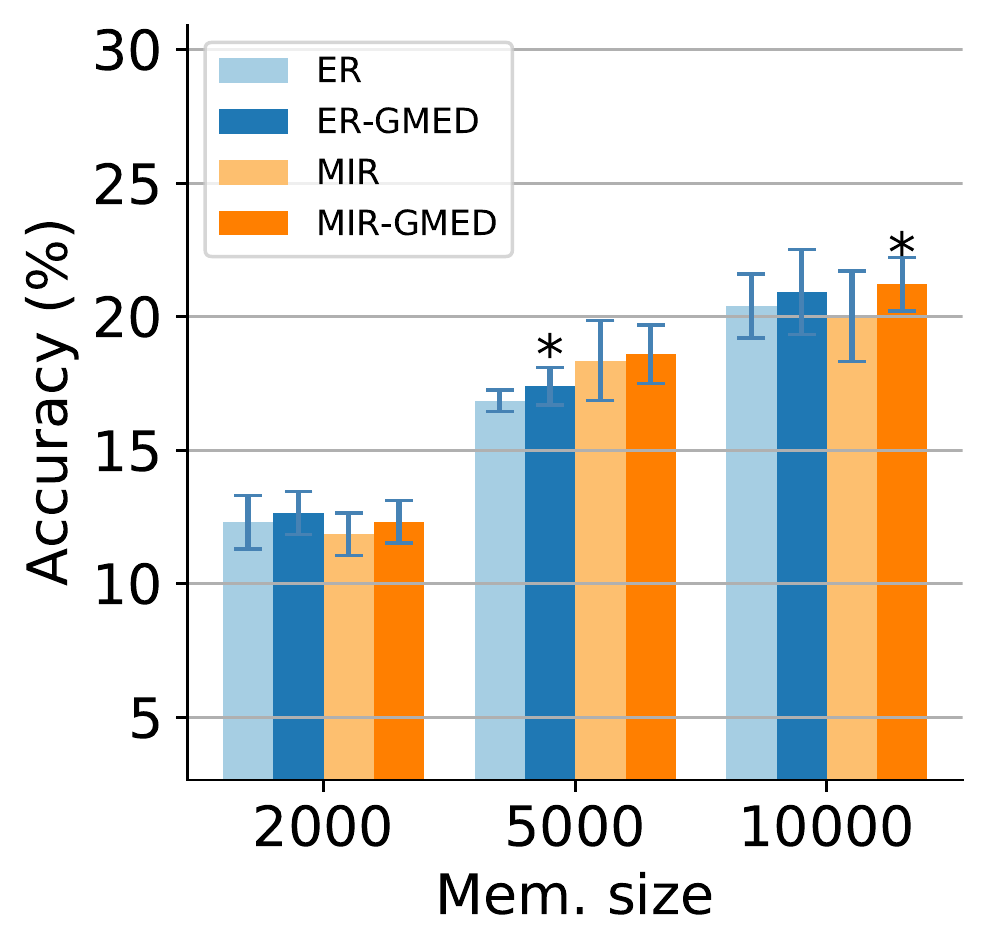}}
    \subfloat[][Split mini-ImgNet]{\includegraphics[width=0.20\textwidth]{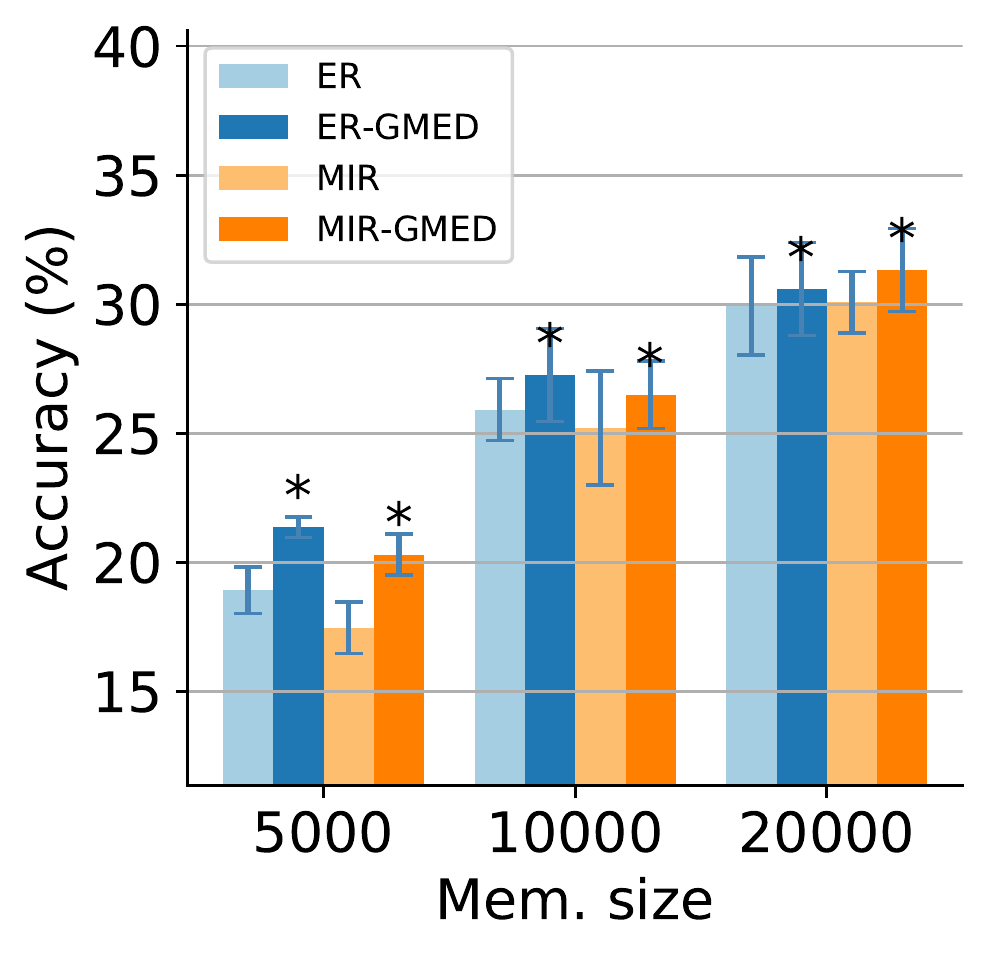}}
    \vspace{0.1cm}
    \caption{\textbf{Performance of ER, GMED$+$ER, MIR, and GMED$+$MIR with different memory sizes}. For mini-ImageNet dataset, we use memory sizes of $1K$, $5K$, $10K$, and $20K$ examples; for Split CIFAR-100 dataset, we use $1K$, $2K$, $5K$ and $10K$; for other datasets, we use $100$, $200$, $500$, and $1000$. $*$ indicates whether the improvement of ER+GMED or MIR+GMED is significant with $p < 0.05$.
    }
    \label{fig:mem_changes}
    \vspace{-0.3cm}
\end{figure*}

\noindent
\textbf{Comparison across CL methods.}
From Table~\ref{tab:main_tab}, we find MIR+GMED achieves the best performance on Split MNIST, Permuted MNIST, and Split CIFAR-100 datasets; while on Rotated MNIST, Split CIFAR-10 and Split mini-ImageNet dataset, ER$_{aug}$+GMED achieves the best performance. Performance of the best performing GMED method could significantly improve over previous SOTA on five datasets.

We further compare with an non-memory based CL approach, CN-DPM~\citep{Lee2020A}, which employs a generative model, a dynamically expanding classifier, and  utilizes a short-term memory (STM). Following the setup in~\citep{hsu2018re}, we set the memory size for GMED so that two methods introduces the same amount of the overhead. Table~\ref{tab:comp_cndpm} shows the results of ER, ER+GMED and the reported results of CN-DPM. Interestingly, ER by itself achieves comparable performance to model expansion approaches. 
ER+GMED further outperforms CN-DPM without any extra memory overhead compared to ER. Similarly, GMED outperforms \textit{task-aware} model expansion approaches such as Prog. NN and the recently proposed compositional model expansion (CompCL) with a smaller memory overhead.

\begin{wraptable}{R}{0.5\textwidth}
\vspace{-0.2cm}
\centering
\caption{\small \textbf{Performance of methods over data streams with fuzzy task boundaries}. In this setup, examples from the next tasks are introduced and gradually dominate the stream when half of the examples from the current task is visited. * indicates whether the improvement is significant ($p < 0.05$)
}
\vspace{-0.1cm}
\label{tab:fuzzy_half}
\scalebox{0.65}{

\begin{tabular}{@{}lcccc@{}}
\toprule
\textbf{Methods / Datasets} & \textbf{\begin{tabular}[c]{@{}c@{}}Split\\ MNIST\end{tabular}} &  \textbf{\begin{tabular}[c]{@{}c@{}}Split\\ CIFAR-10\end{tabular}} &  \textbf{\begin{tabular}[c]{@{}c@{}}Split\\ mini-ImageNet\end{tabular}} \\ \midrule
\textbf{Vanilla}          &      21.53 $\pm$ 0.1                                 &        20.69 $\pm$ 2.4       &                 3.05 $\pm$ 0.6                            \\
\textbf{ER}          &       79.74 $\pm$ 4.0                                    &        37.15 $\pm$ 1.6        &                 26.47 $\pm$ 2.3                            \\
\textbf{MIR}         &              84.80 $\pm$ 1.9                                 &         38.70 $\pm$ 1.7           &            25.83 $\pm$ 1.5                                                \\
\textbf{ER$_{aug}$}  &                  81.30 $\pm$ 2.0                            &              47.97 $\pm$ 3.5          &           30.75 $\pm$ 1.0                     \\ \midrule
\textbf{ER + GMED}   &      82.73$^*$ $\pm$ 2.6                                       &        40.57$^*$ $\pm$ 1.7        &               28.20$^*$ $\pm$ 0.6                   \\                     

\textbf{MIR + GMED}  &                  \textbf{86.17 $\pm$ 1.7}                          &        41.22$^*$ $\pm$ 1.1            &           26.86$^*$ $\pm$ 0.7                     \\ 

\textbf{ER$_{aug}$ + GMED}  &                  82.39$^*$ $\pm$ 3.7                         &              \textbf{51.38$^*$ $\pm$ 2.2}             &           \textbf{31.83 $\pm$ 0.8}                     \\ \bottomrule

\end{tabular}

}

\vspace{-0.2cm}
\end{wraptable}

\noindent \textbf{Performance under Various Memory Sizes.} Figure~\ref{fig:mem_changes} shows the performance of ER, ER+GMED, MIR, and MIR+GMED under various memory sizes. 
The improvement on Split MNIST and Split mini-ImageNet are significant with $p<0.05$ over all memory size setups. 
On Rotated MNSIT and Split CIFAR-10 the improvements are also mostly significant. The improvement on Split CIFAR-100 is less competitive, probably because the dataset is overly difficult for class-incremental learning, from the accuracy around or less than $20\%$ in all setups.

\noindent \textbf{Performance on Data Streams with Fuzzy Task Boundaries}. 
The experiments in Table~\ref{tab:main_tab} assume a clear task boundary.
In Table~\ref{tab:fuzzy_half}, we report the result of using data-streams with fuzzy task boundaries on four datasets. 
We leave the complete results in Table~\ref{tab:fuzzy_full} in Appendix. In this setup, the probability density of a new task grows linearly starting from the point where 50\% of examples of the current task are visited. In particular, GMED improves performance in three datasets across ER, MIR and ER$_{aug}$.


\noindent \textbf{Effectiveness of GMED in Task-Aware Setting.} We additionally compare performance of other task-aware approaches (HAL, GEM, GEM+GMED) in Appendix~\ref{apdx:task_aware}.


\vspace{-0.1cm}
\subsection{Ablation to Study the Effect of Memory-Editing}
\label{ssec:gmed_abl_optimal}
\vspace{-0.2cm}


We present a set of experiments to validate that the gains obtained through the memory-editing step of GMED are indeed helpful towards alleviating catastrophic forgetting.
Further, we show that these gains are distinct from those obtained through random perturbations or simple regularization effects.





\begin{figure}
\vspace{-0.5cm}
\centering
\begin{minipage}[t]{0.43\textwidth}
\centering
\includegraphics[width=\linewidth]{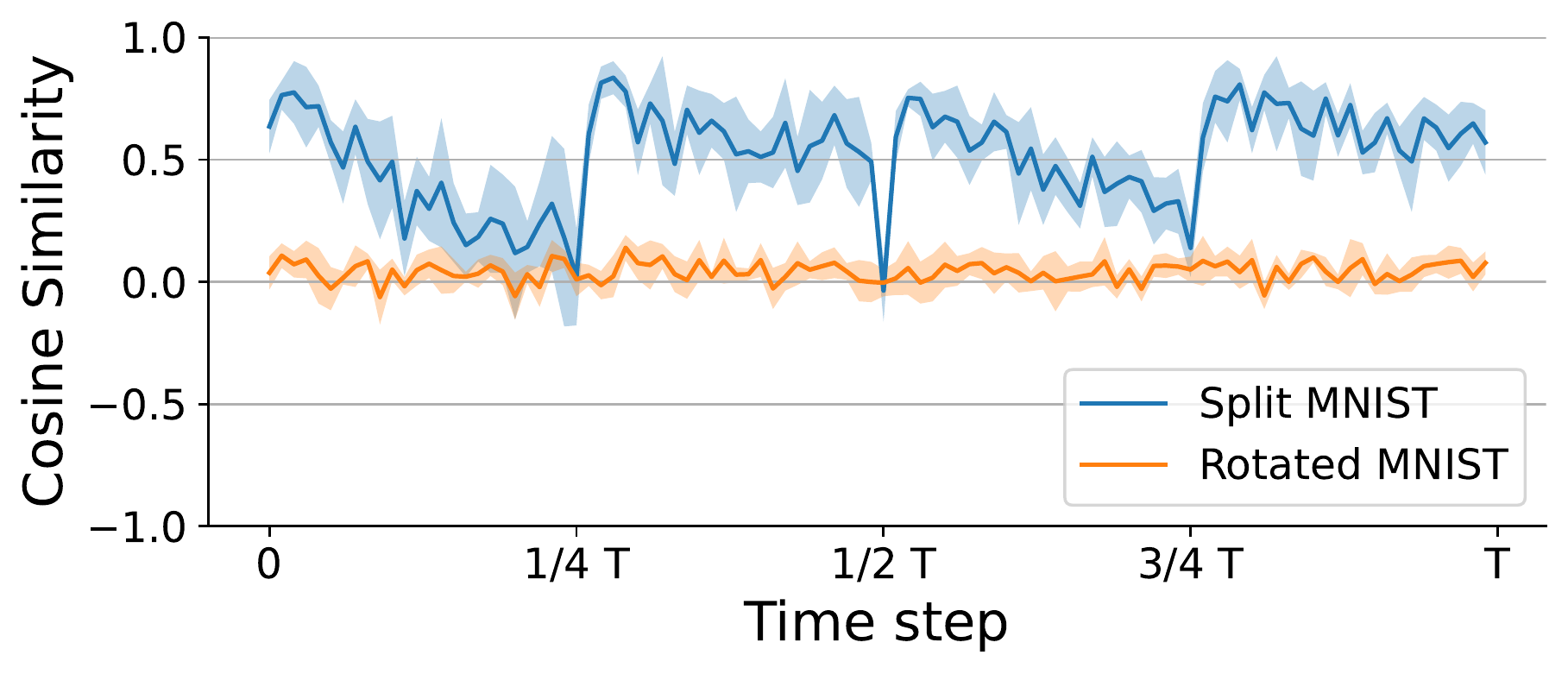}
\caption{\small Cosine similarity between the optimal editing direction and direction from ER+GMED over training. The averaged similarity is $0.523 \pm 0.014$ and $0.035 \pm 0.009$.}
\label{fig:cosine_sim}
\end{minipage}
\hspace{0.3cm}
\begin{minipage}[t]{0.53\textwidth}
\centering
\includegraphics[width=\linewidth]{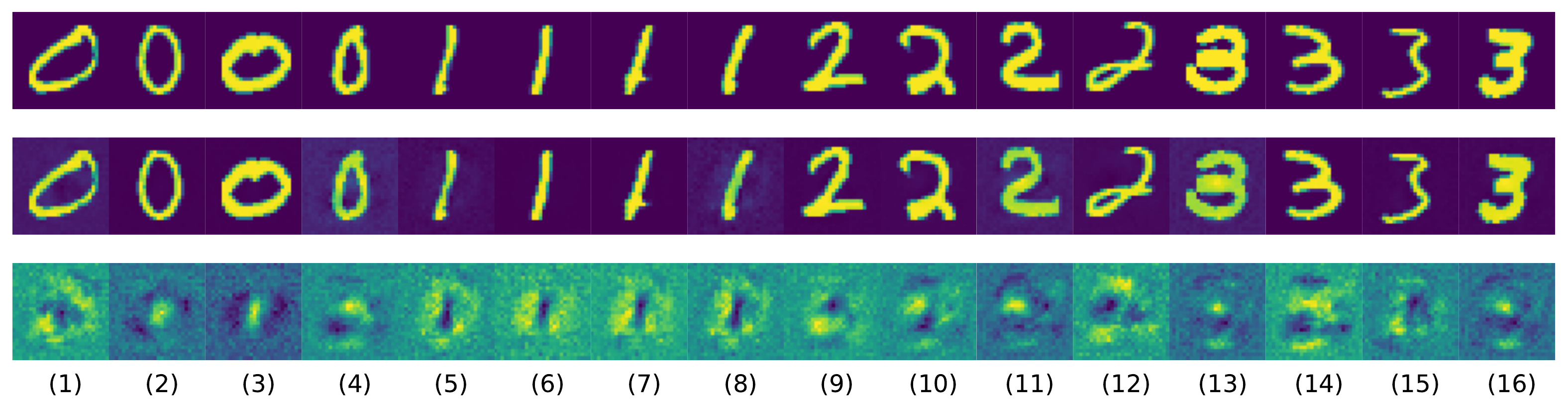}
\caption{\small Visualization of the edited examples in Split MNIST in the ER-GMED approach. The first two rows show examples before and after editing, and the third row shows the differences.}
\label{fig:edit_viz}
\end{minipage}
\vspace{-0.3cm}
\end{figure}

\noindent \textbf{Comparison to Optimal Editing Direction.} 
At a particular time-step $t$, GMED identifies the edit direction using the stream example $(x_D, y_D)$ and the memory sample $(x_m, y_m)$. 
To validate whether the edits proposed by GMED are indeed helpful, we compare GMED-edit to an ``Optimal Edit''  that minimizes the loss increase  (``interference'') of all early training examples in one future time step. 

To compute this Optimal Edit, we note that the total loss increase over all previously encountered examples would be $d_{t:t+1}^{1:t} = \sum_{i=1}^{t} d_{t:t+1}(x_i, y_i) = \sum_{i=1}^{t}  [\ell(x_i, y_i; \theta_{t+1}) - \ell(x_i, y_i; \theta_t)]$, where $\theta_{t+1} = \theta_{t} - \nabla_{\theta}\ell(x_m, y_m; \theta_t) - \nabla_{\theta}\ell(x_D, y_D; \theta_t) $ is the model parameters after the training update on $(x_D, y_D)$ and $(x_m, y_m)$. 
The Optimal Edit direction for the memory example $x_m$ would be the gradient of $d_{t:t+1}^{1:t}$ w.r.t. $x_m$. 
Computing such optimal edits requires access to early training examples and is not practical in an online continual learning setup; we present it only for ablative purposes. 

Figure~\ref{fig:cosine_sim} shows the cosine similarity of update directions (\textit{i.e.}, the gradient of $x_m$) between GMED and the optimal editing strategy over Split MNIST and Rotated MNIST datasets. The averaged similarity over time is $0.523 \pm 0.014$, $0.035 \pm 0.009$ on two datasets, averaged across $10$ runs. Recall that random editing has an expectation of zero similarity. 
On Split MNIST dataset, where the improvement is the most significant, we notice a high similarity between the update directions of GMED and optimal editing. On Rotated MNIST where the improvement is less significant, the similarity is still positive on average but is lower than Split MNIST. It implies GMED-edits are generally consistent with explicitly reducing forgetting, but there is still space to improve. The results also imply the whether the edits of GMED aligns with editing is highly dependent on certain properties of datasets. We further include the classification performance of optimal editing in Appendix~\ref{apdx:optimal_editing}.

\begin{wraptable}{R}{0.49\textwidth}
\vspace{-0.2cm}
\centering
\caption{\small \textbf{Alternative memory editing objectives or simply increasing the number of replayed examples}. Comparison of Random Edit, Adversarial Edit (Adv. Edit) to our proposed objective in GMED.
$^*$ indicates significant improvement ($p{<}0.05$) compared to adversarial edit.\\
}
\vspace{-0.4cm}
\label{tab:ablation}
\scalebox{0.61}{

\begin{tabular}{@{}lcccc@{}}
\toprule
\textbf{Methods / Datasets}  & \textbf{\begin{tabular}[c]{@{}c@{}}Rotated\\ MNIST\end{tabular}} & \textbf{\begin{tabular}[c]{@{}c@{}}Split\\ CIFAR-10\end{tabular}} & \textbf{\begin{tabular}[c]{@{}c@{}}Split\\ mini-ImageNet\end{tabular}} \\ \midrule
\textbf{ER}          & 76.71 $\pm$ 1.6                                                  & 33.30 $\pm$ 3.9                    & 25.92 $\pm$ 1.2                                                        \\                   
\textbf{ER + Random Edit}                   & 76.42 $\pm$ 1.3                                                  & 32.26 $\pm$ 1.9                    & 26.50 $\pm$ 1.3                                                        \\                   
\textbf{ER + Adv. Edit} & 76.13 $\pm$ 1.5 & 31.69 $\pm$ 0.8 & 26.09 $\pm$ 1.6 \\
\textbf{ER + GMED}                                                                                  & \best{77.50 $\pm$ 1.6}                                & \best{34.84$^*$ $\pm$ 2.2}                                     & \best{27.27$^*$ $\pm$ 1.8}                                                       \\
\midrule
\textbf{MIR+Extra 1 batch} &                                       78.07 $\pm$  1.1                                      &              33.81 $\pm$ 2.3                                 &                             24.94 $\pm$ 1.5                                \\
\textbf{MIR+Extra 2 batches} &                             77.10 $\pm$  1.4                                      &              32.36 $\pm$ 2.8                                 &                             24.65 $\pm$ 1.5                                \\

\midrule
\textbf{MIR}                        & 77.50 $\pm$ 1.6                                                  & 34.42 $\pm$ 2.4                    & 25.21 $\pm$ 2.2                                                        \\        
\textbf{MIR + Random Edit}                                                               & 77.19 $\pm$ 1.0                                                  & 35.39 $\pm$ 3.0                              & 24.86 $\pm$ 0.7 \\
\textbf{MIR + Adv. Edit}                                                               & 78.06 $\pm$ 1.5                                                  & 35.79 $\pm$ 0.4                              & 25.48 $\pm$ 1.3      \\
\textbf{MIR + GMED}                                                 & 
\best{79.08$^*$ $\pm$ 0.8}                                                  & \best{36.17 $\pm$ 2.5}                                       & \best{26.29$^*$ $\pm$ 1.2}         \\  
\bottomrule

\end{tabular}

}
\vspace{-0.3cm}
\end{wraptable}

\noindent \textbf{Comparison with alternative Editing Objectives}. 
While GMED objective editing correlates with that of Optimal Edit, we further validate the choice of the objective function and consider two alternatives to the proposed editing objective (Eq.~\ref{eq:editing}): 
(i) Random Edit: memory examples are updated in a random direction with a fixed stride; 
(ii) Adversarial Edit: following the literature of adversarial example construction~\citep{Goodfellow2015ExplainingAH}, the edit increases the loss of memory by following the gradient $
\textrm{sgn}\;\nabla_x \ell(x_m, y_m; \theta)$.
We report the comparison in Table~\ref{tab:ablation}.

We notice ER+Random Edit outperforms ER on split mini-ImageNet, which indicates adding random noise to memory examples helps in regularization.
Even so, GMED-edits are as good or outperforms both random and adversarial edits across multiple datasets and models. This validates our choice of the editing objective.

\noindent \textbf{Comparison to Increasing the Number of Replayed Examples.} Recall that in MIR-GMED, we sample two independent subset of memory examples to perform editing and replay (Sec.~\ref{ssec:gmed_mir}). 
For a fairer comparison, we replay one (or more) extra subset of examples. 
Table~\ref{tab:ablation} shows the performance does not improve as replaying more examples. 
We hypothesize that the performance is bottle-necked by the size of the memory and not the number of examples replayed.

\begin{figure*}
\vspace{-0.5cm}
    \centering
    \subfloat[][{\scriptsize Split MNIST}]{\includegraphics[width=0.16\textwidth]{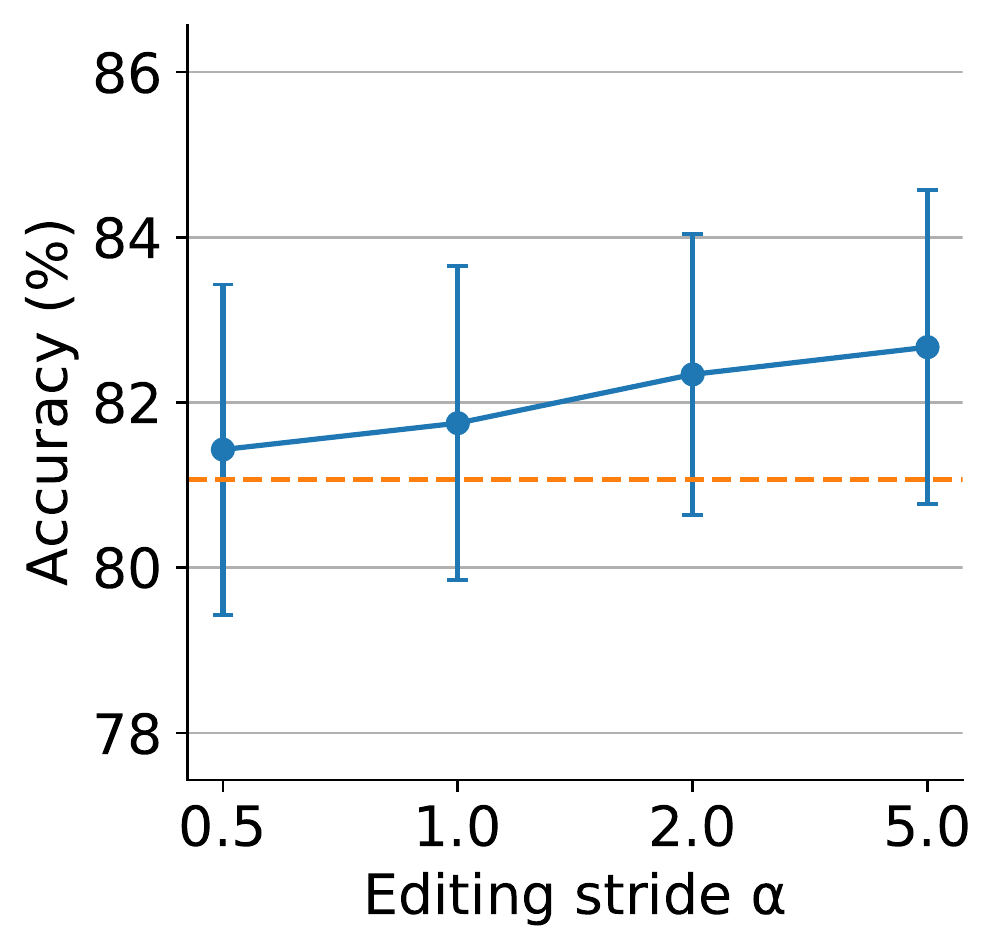}}
    \subfloat[][{\scriptsize Split CIFAR-10}]{\includegraphics[width=0.16\textwidth]{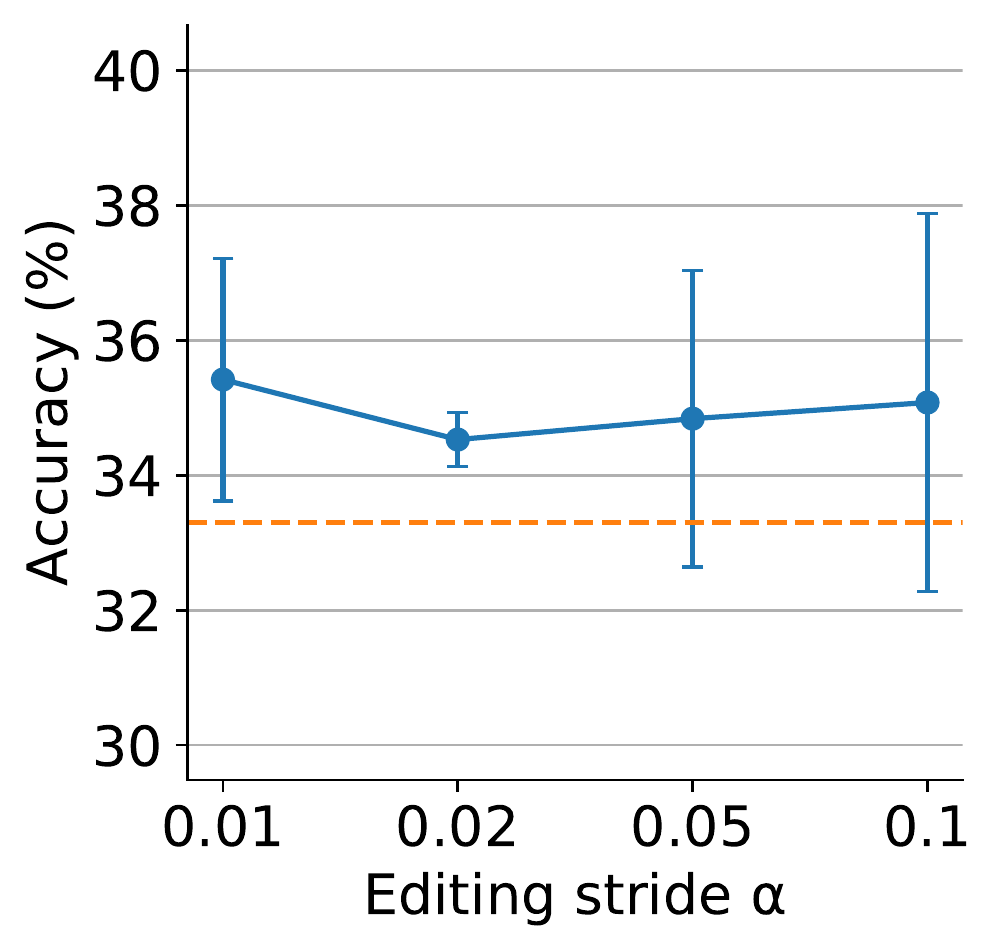}}
    \subfloat[][{\scriptsize Split mini-ImgNet}]{\includegraphics[width=0.16\textwidth]{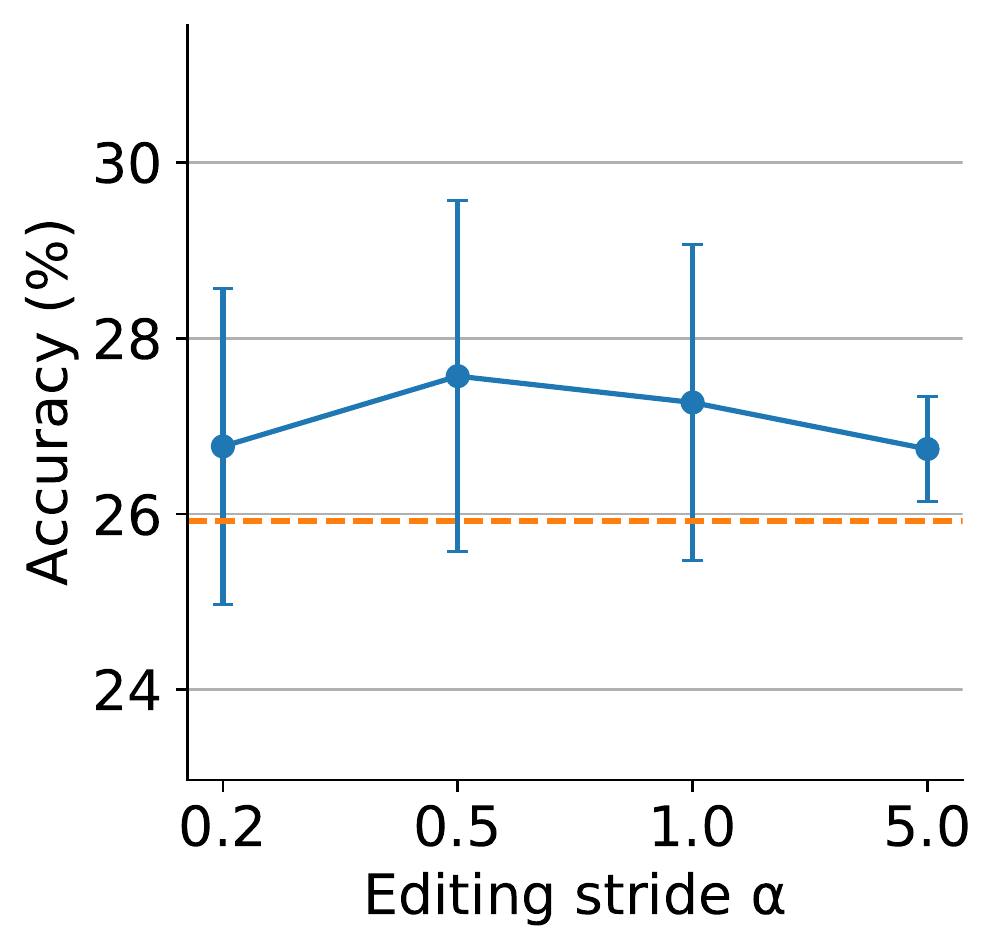}}
    \subfloat[][{\scriptsize Split MNIST} ]{\includegraphics[width=0.16\textwidth]{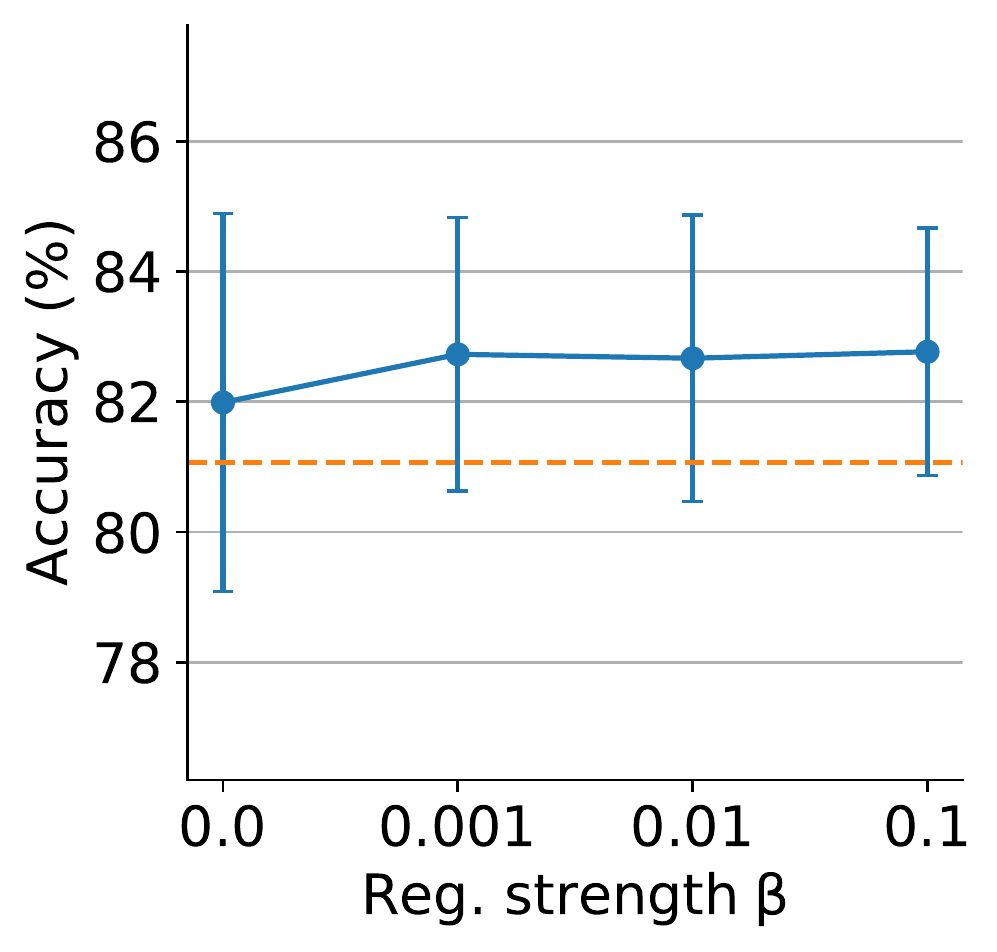}}
    \subfloat[][{\scriptsize Split CIFAR-10}]{\includegraphics[width=0.16\textwidth]{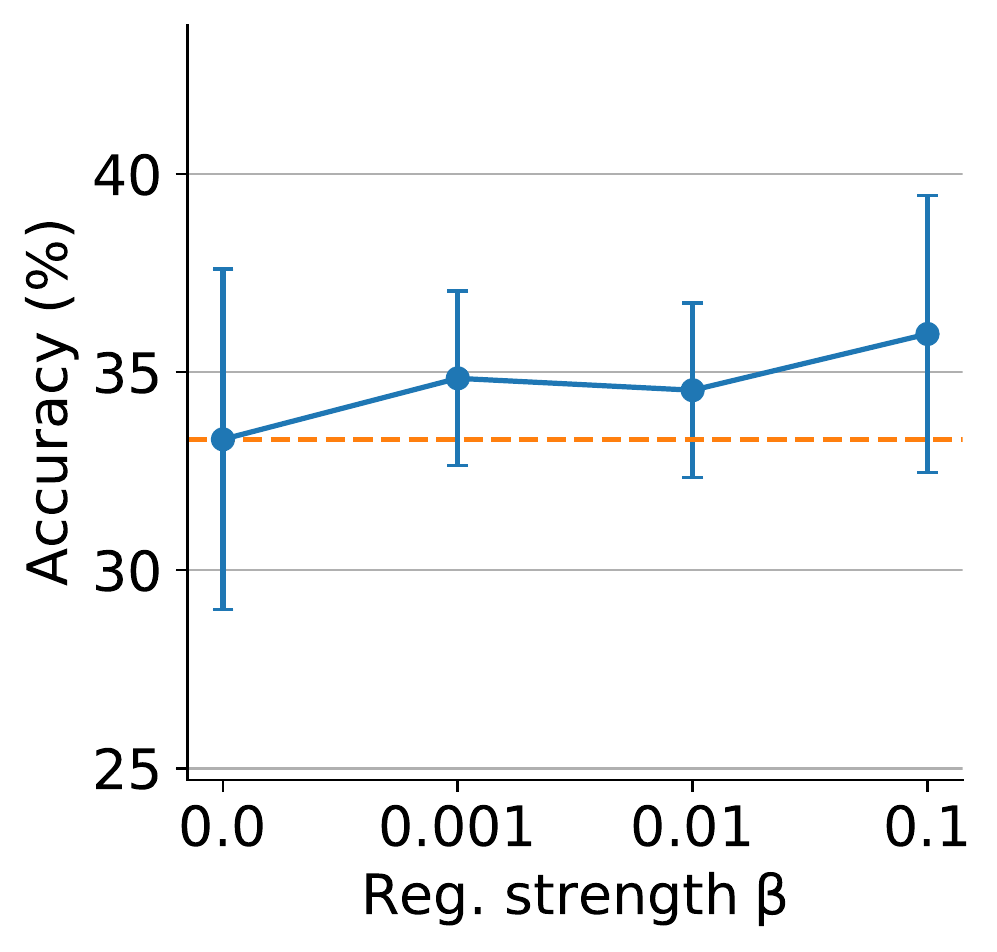}}
    \subfloat[][{\scriptsize Split mini-ImgNet}]{\includegraphics[width=0.16\textwidth]{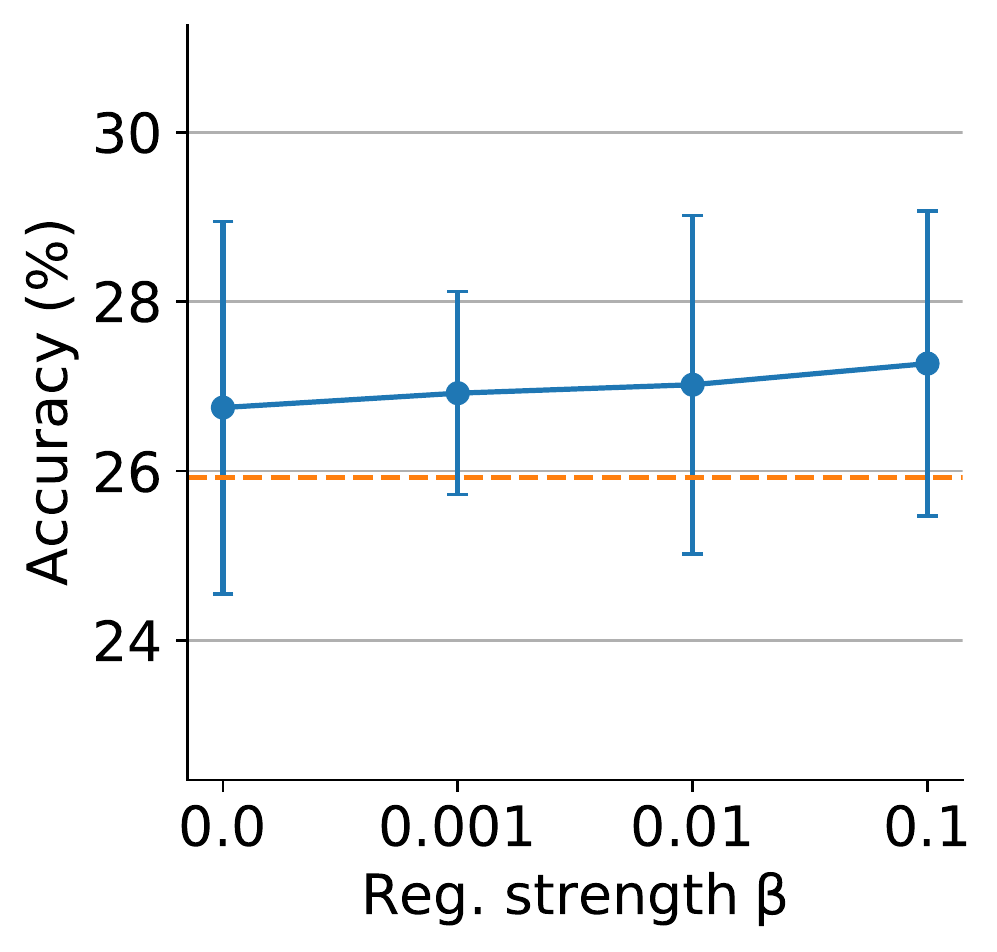}}
    \vspace{-0.0cm}
    \caption{\textbf{Parameter sensitivity analysis} of the editing stride $\alpha$ (a,b,c) and the regularization strength $\beta$ (d,e,f) for ER+GMED across various datasets. The dashed horizontal line indicates the performance of ER.  
    }
    \label{fig:hyp_sensitivity}
    \vspace{-0.2cm}
\end{figure*}

\noindent \textbf{Case study on Edited Memory Examples.}
In Figure~\ref{fig:edit_viz}, we show visualizations of editing of memory examples. 
The examples are drawn from first two task ($0/1$, $2/3$) in the Split MNIST dataset using ER+GMED. 
The first and second row show the original and edited examples, 
noted as $x_{\textrm{before}}$ and $x_{\textrm{after}}$ respectively.
The third row shows the difference between two $\Delta x = x_\textrm{after} - x_{\textrm{before}}$.
While there is no significant visual differences between original and edited examples, in the difference $\Delta x$, we find examples with  exaggerated contours (\eg{} examples (1) and (12)) and blurring (\eg{} examples (2), (3), (5), and (6)). 
Intuitively, ambiguous examples are exaggerated while typical examples are blurred.
Our visualizations supports this intuition: examples (1) and (12) are not typically written digits, while examples (2), (3), (5), and (6) are typical. Appendix \ref{apdx:tsne} provides a t-SNE~\citep{maaten2008visualizing} plot of the edit vector.

\vspace{-0.1cm}
\subsection{Analysis on the GMED Framework}
\label{ssec:gmed_abl_params}
\vspace{-0.2cm}
Here, we analyze and ablate the various components used in the GMED framework and their effect on the final performance.

\noindent \textbf{Hyper-parameter Sensitivity}. 
In Figure~\ref{fig:hyp_sensitivity} we plot the sensitivity of the performance of GMED with respect to two hyper-parameters: the editing stride $\alpha$ and the regularization strength $\beta$ (Eq.~\ref{eq:editing}). 
Clearly, ER+GMED outperforms ER over a broad range of $\alpha$ and $\beta$ setups.
Furthermore, in Figure~\ref{fig:hyp_sensitivity} (d,e,f), better performance for non-zero $\beta$ confirms the benefit of the regularization term. Recall that in our main experiments we tune the hyper-parameters $\alpha$ and $\beta$ with only first three tasks; we note the chosen hyperparameters improve the performance despite they are not always optimal ones.



\textbf{Computational Efficiency.} 
We analyze the \textit{additional} forward and backward computation required by ER+GMED and MIR. Compared to ER, ER+GMED adds $3$ forward and $1$ backward passes to estimate loss increase, and $1$ backward pass to update the example. In comparison, 
MIR adds $3$ forward and $1$ backward passes with $2$ of the forward passes are over a larger set of retrieval candidates. In our experiments, we found GMED has similar training time cost as MIR. 
In Appendix~\ref{apdx:algo_details_mir_gem}, we report the wall-clock time, and observe the run-time of ER+GMED is $1.5$ times of ER.

\begin{wraptable}{R}{0.55\textwidth}
\vspace{-0.2cm}
\centering
\caption{\textbf{Editing examples for multiple gradient steps in ER+GMED.} We tune the editing stride ($\alpha$) using the first three tasks as described in Sec. ~\ref{ssec:gmed_with_er}.
}
\label{tab:more_steps}

\scalebox{0.62}{
\begin{tabular}{@{}lcccc@{}}
\toprule
\textbf{Methods / Datasets} & \textbf{\begin{tabular}[c]{@{}c@{}}Split\\ MNIST\end{tabular}} & \textbf{\begin{tabular}[c]{@{}c@{}}Rotated\\ MNIST\end{tabular}} & \textbf{\begin{tabular}[c]{@{}c@{}}Split\\ CIFAR-10\end{tabular}} & \textbf{\begin{tabular}[c]{@{}c@{}}Split\\ mini-ImageNet\end{tabular}} \\ \midrule
\textbf{1-step Edit}   & 82.21 $\pm$ 2.9                             & 77.50 $\pm$ 1.6                                                  & 34.84 $\pm$ 2.2                    & 27.27 $\pm$ 1.8                                                        \\                   
\textbf{3-step Edit}    & 82.55 $\pm$ 1.9  & 77.37 $\pm$ 1.7 & 34.93 $\pm$ 1.4 & \best{27.36 $\pm$ 1.7} \\
\textbf{5-step Edit}    & \best{83.11 $\pm$ 1.9}  & \best{77.53 $\pm$ 1.6} & \best{36.82 $\pm$ 1.8} & 
26.36 $\pm$ 2.0 \\
\bottomrule
\end{tabular}
}
\vspace{-0.2cm}
\end{wraptable}

\noindent \textbf{Increasing Steps of Editing.}
 For experiments in Table ~\ref{tab:main_tab}, we performed one editing step over the sampled memory example $x_m$ at time step $t$. In general, we can increase the number of editing steps. The direction for the edit is computed at each step, which makes the approach different from increasing the editing stride ($\alpha$). 
 Table ~\ref{tab:more_steps} indicates that 3-step and 5-step edits in general don't lead to significant improvement in performance while incurring additional computational cost. As such, we use only $1$-edit step across all our experiments. 




\vspace{-0.1cm}
\section{Conclusion}
\label{sec:conclusion}
\vspace{-0.2cm}
In this paper, we propose Gradient based Memory Editing (GMED), a modular framework for memory-based task-free continual learning where examples stored in the memory can be edited.
Importantly, memory examples edited by GMED are encouraged to remain in-distribution but yield increased loss in the upcoming model updates, and thus are more effective at alleviating catastrophic forgetting.
We find that combining GMED with existing memory-based CL approaches leads to consistent improvements across multiple benchmark datasets, only inducing a small computation overhead.
Finally, we perform a thorough ablative study to validate that the gains obtained by GMED can indeed be attributed to its editing operation and careful choice of editing objective.


\section*{Acknowledgements}
This research is supported in part by the Office of the Director of National Intelligence (ODNI), Intelligence Advanced Research Projects Activity (IARPA), via Contract No. 2019-19051600007, the DARPA MCS program under Contract No. N660011924033, the Defense Advanced Research Projects Agency with award W911NF-19-20271, NSF IIS 2048211, NSF SMA 1829268, and gift awards from Google, Amazon, JP Morgan and Sony. 
We would like to thank all the collaborators in USC INK research lab for their constructive feedback on the work.

\bibliography{GMED}
\bibliographystyle{plain2}
\newpage
\appendix

\section{Implementation Details of the Compared Methods}
\label{sec:details_of_baselines}
We included detailed descriptions, and implementation details of some selected baselines in this section.


\noindent \textbullet~~\textbf{Experience Replay (ER)}~\citep{Robins1995CatastrophicFR, Rolnick2019ExperienceRF} stores examples in a fix-sized memory for future replay. We use reservoir sampling to decide which examples to store and replace. Following prior works~\citep{Aljundi2018TaskFreeCL, Aljundi2019OnlineCL, Chaudhry2020UsingHT}, at each time step we draw the same number of examples as the batch size from the memory to replay, which are both set to $10$. The algorithm is applied as is to the task-free scenario. 

\noindent \textbullet~~\textbf{Gradient Episodic Memory (GEM)}~\citep{LopezPaz2017GradientEM} also stores examples in a memory. Before each model parameter update, GEM projects the gradient of model parameters so that the update does not incur loss increase on any previous task. The approach is however not task-free. We used a memory strength $\gamma=0.5$ for Split MNIST and Rotated MNIST following the original work and $\gamma=0.5$ following~\citep{Aljundi2019GradientBS}. We applied the same $\gamma$ for remaining datasets.

\noindent \textbullet~~\textbf{Averaged Gradient Episodic Memory (AGEM)} \citep{chaudhry2018efficient} prevents the average loss increase on a randomly drawn subset of examples from the memory. 
We draw $k=256$ examples to compute the regularization at each iteration. Note than this setup of $k$ is much larger than the number of examples drawn for replay ($k=10$) in ER approaches. The approach is task-free.

\noindent \textbullet~~\textbf{Bayesian Gradient Descent (BGD)}~\citep{Zeno2018TaskAC} is a regularization-based continual learning algorithm. It adjusts the learning rate for parameters by estimating their certainty, noting their importance to previous data. We tune the initial standard deviation of parameters and set it as 0.011 for Split CIFAR-10 and Split CIFAR-100, 0.05 for permuted MNIST, and 0.017 for Split MNIST and Rotated MNIST. We also tune the ``learning rate'' hyperparameter $\eta$ and set it as 8 for Split CIFAR-10 and Split CIFAR-100, and 1 for Permuted MNIST, Split MNIST, and Rotated MNIST. The approach is task-free. 

\noindent \textbullet~~\textbf{Gradient based Sample Selection (GSS)}~\citep{Aljundi2019GradientBS} builds upon Experience Replay (ER) by encouraging the diversity in the stored examples. 
We use GSS-Greedy, which is the best performing variant. The approach is task-free.

\noindent \textbullet~~\textbf{Hindsight Anchor Learning (HAL)}~\citep{Chaudhry2020UsingHT} learns a pseudo ``anchor'' example per task per class in addition to the replay memory by maximizing its estimated forgetting, and tries to fix model outputs on the anchors at training. However, unlike GMED, HAL estimates forgetting by comparing the loss before and after the model performs updates with the replay memory examples (and thus forgetting is estimated with ``hindsight''). We refer to hyperparameters in the original work and set the mean embedding strength $\gamma=0.1$, anchor learning rate as 0.001, gradient steps on anchor $k$ as 100, and fine-tune 50 epochs on the memory to estimate forgetting across datasets. The approach is not task-free. 

\noindent \textbullet~~\textbf{Maximally Interfering Retrieval (MIR)}~\citep{Aljundi2019OnlineCL} improves ER by selecting top forgettable examples from the memory for replay. Following the official implementation, we evaluate forgetting on a candidate set of $25$ examples for mini-ImageNet dataset, and $50$ examples for others. 
While the approach is task-free, the official implementation filter out memory examples that belong to the same task as the current data stream, which assumes knowledge about tasks boundaries. 
We remove this operation to adapt the method to the task-free setup. 
Therefore, our results are not directly comparable to the official results. 

\noindent \textbullet~~\textbf{Neural Dirichlet Process Model for Continual Learning (CN-DPM)}~\citep{Lee2020A} is a task-free model-expansion based continual learning algorithm. We report the official results in the paper. In the comparison study between ER/ER+GMED with CN-DPM, for the base model in ER/ER+GMED, we use the full expanded model in CN-DPM (\ie, the model architecture when the training ends in CN-DPM). 
We use the same optimizer and the learning rate as CN-DPM for compared methods in this set of experiments. 

\noindent \textbullet~~\textbf{Progressive Neural Networks (Prog. NN)}~\citep{Rusu2016ProgressiveNN} is a task-aware model-expansion base approach. We treat linear layers and ResNet blocks as basic components of expansion in MLP and ResNet-18 models, \ie, the model creates a new linear layer or ResNet block at each layer of the model when it encounters a new task. In addition to added components, it also introduces the connectivity between the current component and the components in previous layers as learnable parameters. The model before expansion has the same architecture and sizes as models used in other approaches such as ER.

\noindent \textbullet~~\textbf{Compositional Continual Learning (CompCL)}~\citep{mendez2021lifelong} is another task-aware model-expansion based approach. When the model encounters a new task, the approach performs two-stage model updates by first learning the architecture over existing and one new model component, and then performing updates on parameters of components. While the original work adds one shared component for all layers, we find it helpful to add separate components for all layers, possibly because the tasks in our setup have low similarity.

\section{Algorithmic and Implementation Details for GMED variants}
\label{apdx:algo_details_mir_gem}

\begin{algorithm}
\begin{small}
\caption{\small Memory Editing with ER$_{aug}$ (ER$_{aug}$+GMED)}
\begin{algorithmic}
   \STATE {\bfseries Input:} learning rate $\tau$, edit stride $\alpha$, regularization strength $\beta$, decay rate $\gamma$, model parameters $\theta$
   \STATE {\bfseries Receives}: stream example $(x_D, y_D)$
   \STATE {\bfseries Initialize}: replay memory $M$
      
    \smallskip
   \FOR{$t=1$ {\bfseries to} $T$}
      \smallskip
      
   \STATE \texttt{\small // draw a random mini-batch for edit and augmentation respectively}
   \STATE $(x_e,y_e) \sim M$
   \STATE $k \leftarrow \textrm{replayed\_time} (x_e,y_e)$
   \STATE $\ell_\textrm{before}\leftarrow \textrm{loss}(x_e,y_e,\theta_t)$
   \STATE $\ell_\textrm{stream}\leftarrow \textrm{loss}(x_D, y_D,\theta_t)$ 
   
   \smallskip
   \STATE \texttt{\small//update model parameters with stream examples, discarded later}
      \smallskip
   \STATE $\theta_t^\prime \leftarrow  \textrm{SGD}(\ell_\textrm{stream}, \theta_t, \tau)$ 

    \smallskip
   \STATE \texttt{\small//evaluate forgetting of memory examples}
   \STATE $\ell_\textrm{after}\leftarrow \textrm{loss}(x_e,y_e,\theta_t^\prime)$
   \STATE $d\leftarrow \ell_\textrm{after} - \ell_\textrm{before}$
   
   \smallskip
   \STATE \texttt{\small//edit memory examples}
   \STATE $x_e^\prime \leftarrow x_e + \gamma^k \alpha \nabla_x (d - \beta \ell_\textrm{before})$
  \STATE replace $(x_e, y_e)$ with $(x_e^\prime, y_e)$ in $M$
   
   \smallskip    
   \STATE $(x_e^a, y_e) \leftarrow \textrm{data\_augmentation}(x_e, y_e)$
   
   \STATE \texttt{\small//replay edited and augmented examples}
   \STATE $\ell = \textrm{loss}(\{(x^a_e,y_e), (x_e^\prime, y_e), (x_D, y_D)\}, \theta_t)$
   \smallskip
   \STATE $\theta_{t+1} \leftarrow \textrm{SGD}(\ell, \theta_t, \tau)$
   
   
   \STATE $\textrm{reservoir\_update}(x_D, y_D, M) $
      \smallskip
   \ENDFOR
\end{algorithmic}
\label{algo:gmed_er_aug}
\end{small}
\end{algorithm}
\begin{algorithm}
\begin{small}
\caption{\small Memory Editing with MIR (MIR+GMED)}
\begin{algorithmic}
   \STATE {\bfseries Input:} learning rate $\tau$, edit stride $\alpha$, regularization strength $\beta$, decay rate $\gamma$, model parameters $\theta$
   \STATE {\bfseries Receives}: stream example $(x_D, y_D)$
   \STATE {\bfseries Initialize}: replay memory $M$
      
    \smallskip
   \FOR{$t=1$ {\bfseries to} $T$}
      \smallskip
      
   \STATE \texttt{\small // draw a random mini-batch for edit}
   \STATE $(x_e,y_e) \sim M$
   \STATE $k \leftarrow \textrm{replayed\_time} (x_e,y_e)$
   \STATE $\ell_\textrm{before}\leftarrow \textrm{loss}(x_e,y_e,\theta_t)$
   \STATE $\ell_\textrm{stream}\leftarrow \textrm{loss}(x_D, y_D,\theta_t)$ 
   
   \smallskip
   \STATE \texttt{\small // retrieve a separate mini-batch of examples for replay with MIR}   
   \STATE $(x_m,y_m) \leftarrow \textrm{MIR-retrieve}(M, \theta)$

   \smallskip
   \STATE \texttt{\small//update model parameters with stream examples, discarded later}
   \STATE $\theta_t^\prime \leftarrow  \textrm{SGD}(\ell_\textrm{stream}, \theta_t, \tau)$ 

    \smallskip
   \STATE \texttt{\small//evaluate forgetting of memory examples}
   \STATE $\ell_\textrm{after}\leftarrow \textrm{loss}(x_e,y_e,\theta_t^\prime)$
   \STATE $d\leftarrow \ell_\textrm{after} - \ell_\textrm{before}$
   
   \smallskip
   \STATE \texttt{\small//edit memory examples}
      \smallskip

   \STATE $x_e^\prime \leftarrow x_e + \gamma^k \alpha \nabla_x (d - \beta \ell_\textrm{before})$
  \STATE replace $(x_e, y_e)$ with $(x_e^\prime, y_e)$ in $M$
   
 \smallskip    
   
   \STATE \texttt{\small//replay examples retrieved by MIR}
   \STATE $\ell = \textrm{loss}(\{(x_m,y_m),(x_D, y_D)\}, \theta_t)$
   \smallskip
   \STATE $\theta_{t+1} \leftarrow \textrm{SGD}(\ell, \theta_t, \tau)$
      \smallskip

   \STATE $\textrm{reservoir\_update}(x_D, y_D, M) $
      \smallskip
   \ENDFOR
\end{algorithmic}
\label{algo:gmed_mir}
\end{small}
\end{algorithm}
\begin{algorithm}
\begin{small}
\caption{\small Memory Editing with GEM (GEM+GMED)}
\begin{algorithmic}
   \STATE {\bfseries Input:} learning rate $\tau$, edit stride $\alpha$, regularization strength $\beta$, decay rate $\gamma$, model parameters $\theta$
   \STATE {\bfseries Receives}: stream example $(x_D, y_D)$
   \STATE {\bfseries Initialize}: replay memory $M$
      
    \smallskip
   \FOR{$t=1$ {\bfseries to} $T$}
      \smallskip
      
   \STATE \texttt{\small // draw a random mini-batch for edit}
   \STATE $(x_e,y_e) \sim M$
   \STATE $k \leftarrow \textrm{replayed\_time} (x_e,y_e)$
   \STATE $\ell_\textrm{before}\leftarrow \textrm{loss}(x_e,y_e,\theta_t)$
   \STATE $\ell_\textrm{stream}\leftarrow \textrm{loss}(x_D, y_D,\theta_t)$ 
   
   \smallskip
   \STATE \texttt{\small//update model parameters with stream examples, discarded later}
      \smallskip
   \STATE $\theta_t^\prime \leftarrow  \textrm{SGD}(\ell_\textrm{stream}, \theta_t, \tau)$ 

    \smallskip
   \STATE \texttt{\small//evaluate forgetting of memory examples}
   \STATE $\ell_\textrm{after}\leftarrow \textrm{loss}(x_e,y_e,\theta_t^\prime)$
   \STATE $d\leftarrow \ell_\textrm{after} - \ell_\textrm{before}$
   
   \smallskip
   \STATE \texttt{\small//edit memory examples}
   \STATE $x_e^\prime \leftarrow x_e + \gamma^k \alpha \nabla_x (d - \beta \ell_\textrm{before})$
  \STATE replace $(x_e, y_e)$ with $(x_e^\prime, y_e)$ in $M$
   
   \smallskip    
 
   \STATE \texttt{\small//Regularized model updates with GEM using the full memory}
   \STATE $\textrm{GEM\_regularized\_update}(x_D, y_D, M, \theta)$
   
   \STATE \texttt{\small//Update the replay memory following GEM}  
   \STATE $\textrm{memory\_update}(x_D, y_D, M) $
      \smallskip
   \ENDFOR
\end{algorithmic}
\label{algo:gmed_gem}
\end{small}
\end{algorithm}

\textbf{Algorithmic Details.}
We present the algorithmic details of ER$_{aug}$+GMED, MIR+GMED and GEM+GMED in Algorithms~\ref{algo:gmed_er_aug},~\ref{algo:gmed_mir} and~\ref{algo:gmed_gem}. The main difference in MIR+GMED and GEM+GMED compared to ER+GMED is that we edit a separate mini-batch of memory examples from the mini-batch used for replay or regularization. 
Similarly, in ER$_{aug}$+GMED, we additionally replay a mini-batch of edited examples after data augmentation. The motivations for such design are discussed in Sec.~\ref{ssec:gmed_mir}.


\textbf{Implementation Details.}
We implemented our models with PyTorch 1.0. We train our models with a single GTX 1080Ti or 2080Ti GPU, and we use CUDA toolkit 10.1. We use a mini-batch size of $10$ throughout experiments for all approaches.

\textbf{Training Time Cost.}
For ER+GMED, training on Split MNIST, Permuted MNIST, and Rotated MNIST takes 7 seconds, 30 seconds, and 31 seconds respectively (excluding data pre-processing). Training on Split CIFAR-10, Split CIFAR-100, and Split mini-ImageNet takes 11 minutes, 14 minutes, and 46 minutes respectively. In comparison, training with ER takes 6 seconds, 23 seconds, 16 seconds, 7 minutes, 10 minutes, and 32 minutes respectively on six datasets.

\paragraph{Experiment Configuration for Comparison to Model-Expansion based Methods.} Model expansion based approaches such as Progressive Networks~\citep{Rusu2016ProgressiveNN}, Compositional Lifelong Learning~\citep{mendez2021lifelong}, and CN-DPM~\citep{Lee2020A} involves additional overhead by introducing new model parameters over time. We compute such overhead as the number of extra parameters when the model has fully expanded. Let $n_e$, $n_b$ be the number of parameters in a fully-expanded model and a regular model in non-expansion based approaches (\eg, ER) respectively, and the difference is $\Delta n = n_e - n_b$. We report such overhead in the equivalent number of training examples that can be stored in a memory: given an input image with dimension $(c,h,w)$ and the overhead $\Delta n$, the equivalent number of examples is computed as $4\Delta n / (chw+1)$. The coefficient 4 is added because storing a model parameters in float32 type takes up 4 times as much memory as a pixel or a label $y$. For CN-DPM, the short-term memory used to store past examples also counts as overhead.

\section{Hyper-parameter Setup}
\label{apdx:hyp_setup}

Throughout experiments other than results reported in Table~\ref{tab:comp_cndpm}, we use SGD optimizer with a learning rate of $0.05$ for MNIST datasets, $0.1$ for Split CIFAR-10 and Split mini-ImageNet datasets, and $0.03$ for the Split CIFAR-100 dataset across all approaches. The learning rate is tuned with ER method and fixed for all other approaches. For results reported in Table~\ref{tab:comp_cndpm}, we use the same optimizer and learning rates as CN-DPM. See Sec.~\ref{sec:details_of_baselines} for setup of method specific hyperparameters. 


\textbf{GMED variants.}
Besides, GMED introduces two additional hyper-parameters: the stride of the editing $\alpha$, and the regularization strength $\beta$. 
As we assume no access to the full data stream in the online learning setup, we cannot select hyper-parameters according to validation performance after training on the full stream. Therefore, 
we tune the hyper-parameters with only the validation set of first three tasks, following the practice in~\citep{chaudhry2018efficient}. 
The tasks used for hyper-parameter search are included for computing the final accuracy, following~\citep{Ebrahimi2020Uncertainty-guided}. 
We perform a grid search over all combinations of $\alpha$ and $\beta$ and select the one with the best validation performance on the first three tasks. 
We select $\alpha$ from  $[0.01, 0.03, 0.05, 0.07, 0.1, 0.5, 1.0, 5.0, 10.0]$, and select $\beta$ from $[0, 10^{-3}, 10^{-2}, 10^{-1}, 1]$. 
We tune two hyper-parameters on ER+GMED and share it across MIR+GMED and GEM+GMED. 
We tune a separate set of hyper-parameters for ER$_{aug}$+GMED. 
Table~\ref{tab:hyp_selection} reports the optimal hyper-parameters selected for each dataset. 

\begin{table*}[h]
\centering
\caption{Hyperparamters of the editing stride and the regularization strength selected for ER+GMED. 
}

\label{tab:hyp_selection}
\scalebox{0.8}{\begin{tabular}{@{}lcc@{}}

\toprule
\textbf{Dataset / Hyper-param}                    & \textbf{\begin{tabular}[c]{@{}c@{}}Editing \\ stride $\alpha$\end{tabular}} & \textbf{\begin{tabular}[c]{@{}c@{}}Regularization \\ strength $\beta$\end{tabular}} \\ \midrule
\rowcolor{na} \textit{ER+GMED, MIR+GMED, GEM+GMED}              &  & \\
Split MNIST         & 5.0                                                                           & 0.01                                                                                  \\
Permuted MNIST      & 0.05                                                                           & 0.001                                                                                 \\
Rotated MNIST       & 1.0                                                                           & 0.01                                                                                  \\
Split CIFAR-10         & 0.05                                                                          & 0.001                                                                                 \\
Split CIFAR-100         & 0.01                                                                          & 0.001                                                                                 \\
Split mini-ImageNet & 1.0                                                                           & 0.1                                                                                   \\  \midrule
\rowcolor{na} \textit{ER$_{aug}$ + GMED}              &  & \\
Split MNIST         & 5.0                                                                           & 0.001                                                                                  \\
Permuted MNIST      & 0.05                                                                           & 0.001                                                                                 \\
Rotated MNIST       & 1.0                                                                           & 0.01                                                                                  \\
Split CIFAR-10         & 0.07                                                                          & 0.01                                                                                 \\
Split CIFAR-100         & 0.05                                                                          & 0.001                                                                                 \\
Split mini-ImageNet &  0.5                                                                          & 0.0                                                                                 \\ 
\bottomrule

\end{tabular}}

\end{table*}


\paragraph{Sensitivity to $\gamma$.}
Table~\ref{tab:sensitivity_gamma} shows the performance of GMED under various decay rates of the editing stride $\gamma$. We find $\gamma=1.0$ (\ie, no decay) consistently outperforms performance when $\gamma$ is less than $1$. It implies it is not necessary to explicitly discourage the deviation of edited examples from original examples with the hyper-parameter $\gamma$.

\begin{table*}[h]
\centering
\caption{Sensitivity of the performance of GMED to the decay rate of the editing stride ($\gamma$).}
\label{tab:sensitivity_gamma}
\scalebox{0.72}{
\begin{tabular}{@{}lcccccc@{}}
\toprule
\textbf{Methods / Datasets} & \textbf{\begin{tabular}[c]{@{}c@{}}Split\\ MNIST\end{tabular}} & \textbf{\begin{tabular}[c]{@{}c@{}}Permuted\\ MNIST\end{tabular}} & \textbf{\begin{tabular}[c]{@{}c@{}}Rotated\\ MNIST\end{tabular}} & \textbf{\begin{tabular}[c]{@{}c@{}}Split\\ CIFAR-10\end{tabular}} & \textbf{\begin{tabular}[c]{@{}c@{}}Split\\ CIFAR-100\end{tabular}} & \textbf{\begin{tabular}[c]{@{}c@{}}Split\\ mini-ImageNet\end{tabular}} \\ \midrule
\textbf{ER} & 81.07 $\pm$ 2.5                                                & 78.65 $\pm$ 0.7                                                   & 76.71 $\pm$ 1.6                                                  & 33.30 $\pm$ 3.9                               &     20.11 $\pm$ 1.2                    & 25.92 $\pm$ 1.2                                                        \\
\textbf{ER + GMED$_{\gamma=0.9}$}   &      82.28 $\pm$ 1.7                            &        79.07 $\pm$ 0.6                       &                    77.22 $\pm$ 1.2                    &        33.85 $\pm$ 1.4        &         20.06 $\pm$ 1.8         &                  26.92 $\pm$ 1.9                   \\                                                   
\textbf{ER + GMED$_{\gamma=0.99}$}     &      82.60 $\pm$ 2.2                 &           79.15 $\pm$ 0.6              &        77.35 $\pm$ 1.3              &      34.10 $\pm$ 3.4           &   19.90 $\pm$ 1.5        &                          27.69 $\pm$ 0.7         \\
\textbf{ER + GMED$_{\gamma=1.0}$}    &  82.67 $\pm$ 1.9                                           & 78.86 $\pm$ 0.7                                                  & 77.09 $\pm$ 1.3                                              & 34.84 $\pm$ 2.2                              &      20.93 $\pm$ 1.6          & 27.27 $\pm$ 1.8                                                    \\
\midrule
\textbf{MIR}  & 85.72 $\pm$ 1.2                                                & 79.13 $\pm$ 0.7                                                   & 77.50 $\pm$ 1.6                                                  & 34.42 $\pm$ 2.4                          & 20.02 $\pm$ 1.7                      & 25.21 $\pm$ 2.2 \\
\textbf{MIR + GMED$_{\gamma=0.9}$}  & 85.67 $\pm$ 2.2                                                & 79.99 $\pm$ 0.7                                                   & 78.45 $\pm$ 1.3                                                  & 34.98 $\pm$ 0.5                          &    19.76 $\pm$ 1.7                      & 25.96 $\pm$ 1.2 \\
\textbf{MIR + GMED$_{\gamma=0.99}$}     &     86.76 $\pm$ 1.2                  &         79.76 $\pm$ 0.7               &         78.61 $\pm$ 0.6                &    35.78 $\pm$ 3.2      &    20.48 $\pm$ 1.7     &  27.70 $\pm$ 1.3  \\
\textbf{MIR + GMED$_{\gamma=1.0}$}     & 86.52 $\pm$ 1.4                                       & 79.25 $\pm$ 0.8                                               & 79.08 $\pm$ 0.8       & 36.17 $\pm$ 2.5       &                            21.22 $\pm$ 1.0            & 26.50 $\pm$ 1.3               \\
\bottomrule

\end{tabular}
}

\vspace{-0.1cm}
\end{table*}


\section{T-SNE Visualization}
\label{apdx:tsne}

\begin{figure}[h]
    \centering
    \includegraphics[width=0.5\linewidth]{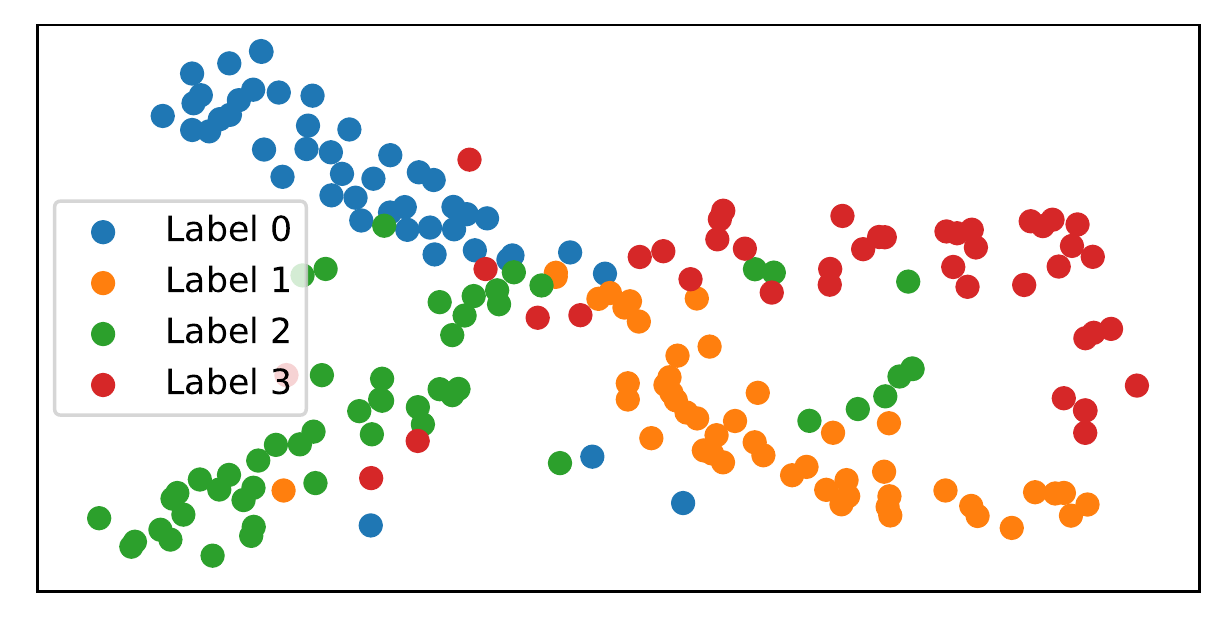}
    \vspace{-0.2cm}
    \caption{A t-SNE visualization of the editing performed on data examples. We use labels from the first two tasks in Split MNIST.}
    \label{fig:edit_tsne}
\end{figure}

In Figure~\ref{fig:edit_tsne}, we show the t-SNE~\citep{maaten2008visualizing} visualization of the editing vector $\Delta x = x_\textrm{after} - x_{\textrm{before}}$ for examples from first 2 tasks in Split MNIST. We note that the editing vectors cluster by the labels of the examples. It implies the editing performed is correlated with the labels and is clearly not random.

\section{Dataset Details}
Following~\citep{Aljundi2019OnlineCL}, we limit the number of training examples per task to 1,000 for all MNIST experiments, including Split MNIST, Permuted MNIST, and Rotated MNIST. The datasets consist of 5, 10, and 20 tasks respectively. In Split CIFAR-10, Split CIFAR-100, and Split mini-ImageNet, each task consists of 10,000, 2,500, and 2,500 training examples. For tuning hyperparameters, we separate out 5\% of the training set examples in the first three tasks as the validation set. For Rotated MNIST, we limit the number of testing examples to 1,000 per task; while for other datasets, we use the complete test sets.

\paragraph{License and Links of Datasets.}
The MNIST dataset can be found in~\url{http://yann.lecun.com/exdb/mnist/}. The dataset is released without a specific license. CIFAR-10 and CIFAR-100 could be found in~\url{https://www.cs.toronto.edu/~kriz/cifar.html}. Similarly, the dataset is released without a license. Mini-ImageNet and its terms of use can be found in~\url{https://mtl.yyliu.net/download/}.

\paragraph{Personally Identifiable Information or Offensive Content in datasets.} Among all datasets we applied, MNIST is a hand-written digit classification dataset, while CIFAR and mini-ImageNet are image classification dataset over common objects. To the best of our knowledge, they do not contain any sensitive information.


\section{Comparison with ER+T}

In addition to data augmentation over the replay memory (ER$_{aug}$),~\citep{Buzzega2020RethinkingER} also studied a number of other tricks, such as exponential LR decay and balanced reservoir sampling. The integrated method, referred to as ER with Tricks (ER+T), achieved SoTA performance in a task-aware, non-online setup over a number of datasets. We adapt ER+T to an online task-free continual learning setup by discarding the tricks not compatible with the online task-free scenario (namely the Bias Control (BiC) trick) and build GMED upon it.
Table~\ref{tab:er_t_gmed} summarizes the results. 
We find ER+T does not outperform ER$_{aug}$, \ie, the tricks other than the data augmentation is not effective in an single-epoch online training setup, except Split MNIST. We further find ER+T+GMED outperforms or performs comparably with ER+T. The improvement is significant on Split MNIST and Split mini-ImageNet datasets.

\begin{table*}[!h]
\centering
\caption{Building GMED over ER+T. $^*$ indicates significant improvement with $p$-value less than 0.05.}
\label{tab:er_t_gmed}
\scalebox{0.72}{
\begin{tabular}{@{}lcccccc@{}}
\toprule
\textbf{Methods / Datasets} & \textbf{\begin{tabular}[c]{@{}c@{}}Split\\ MNIST\end{tabular}} & \textbf{\begin{tabular}[c]{@{}c@{}}Permuted\\ MNIST\end{tabular}} & \textbf{\begin{tabular}[c]{@{}c@{}}Rotated\\ MNIST\end{tabular}} & \textbf{\begin{tabular}[c]{@{}c@{}}Split\\ CIFAR-10\end{tabular}} & \textbf{\begin{tabular}[c]{@{}c@{}}Split\\ CIFAR-100\end{tabular}} & \textbf{\begin{tabular}[c]{@{}c@{}}Split\\ mini-ImageNet\end{tabular}} \\ \midrule
\textbf{ER + T}   &      78.35 $\pm$ 4.5                            &        77.71 $\pm$ 0.7                       &                    \textbf{80.05 $\pm$ 1.3}                  &         \textbf{47.55 $\pm$ 2.6}      &         19.40 $\pm$ 1.5         &                31.25 $\pm$ 1.5                   \\                                                   
\textbf{ER + T + GMED}     &      \textbf{83.02$^{*}$ $\pm$ 0.4}                 &           \textbf{77.92 $\pm$ 0.3}              &      79.96 $\pm$ 0.2               &      47.39 $\pm$ 5.0        &   \textbf{19.75 $\pm$ 1.2}        &                     \textbf{31.84$^{*}$ $\pm$ 1.3} \\ \bottomrule

\end{tabular}
}

\vspace{-0.1cm}
\end{table*}


\section{Full Results of Experiments with Fuzzy Task Boundaries}
\begin{table*}
\centering
\caption{Performance of methods over data streams with fuzzy task boundaries over all six datasets. $^*$ indicates significant improvement with $p$-value less than 0.05.}
\scalebox{0.72}{
\begin{tabular}{@{}lcccccc@{}}
\toprule
\textbf{Methods / Datasets} & \textbf{\begin{tabular}[c]{@{}c@{}}Split\\ MNIST\end{tabular}} & \textbf{\begin{tabular}[c]{@{}c@{}}Permuted\\ MNIST\end{tabular}} & \textbf{\begin{tabular}[c]{@{}c@{}}Rotated\\ MNIST\end{tabular}} & \textbf{\begin{tabular}[c]{@{}c@{}}Split\\ CIFAR-10\end{tabular}} & \textbf{\begin{tabular}[c]{@{}c@{}}Split\\ CIFAR-100\end{tabular}} & \textbf{\begin{tabular}[c]{@{}c@{}}Split\\ mini-ImageNet\end{tabular}} \\ \midrule
\textbf{ER}          &       79.74 $\pm$ 4.0                         &          \textbf{78.98 $\pm$ 0.5}                      &              76.45 $\pm$ 1.2                    &        37.15 $\pm$ 1.6        &           \textbf{21.99 $\pm$ 1.1}                 &                  26.47 $\pm$ 2.3                            \\
\textbf{ER + GMED}   &      \textbf{82.73$^{*}$ $\pm$ 2.6}                            &        78.91 $\pm$ 0.5                       &                    \textbf{76.55 $\pm$ 1.0}                    &        \textbf{40.57$^{*}$ $\pm$ 1.7}        &         21.79 $\pm$ 1.9         &                  \textbf{28.20$^*$ $\pm$ 0.6}                   \\                                        \midrule
\textbf{MIR}         &              85.80 $\pm$ 1.9                           &               \textbf{79.31 $\pm$ 0.7}                            &         77.04 $\pm$ 1.2                                &         38.70 $\pm$ 1.7           &       21.57 $\pm$ 1.4            &           25.83 $\pm$ 1.5                                                \\
\textbf{MIR + GMED}  &                  \textbf{86.17 $\pm$ 1.7}                  &                  79.26 $\pm$ 0.8                   &             \textbf{77.56$^*$ $\pm$ 1.1}                          &              \textbf{41.22$^*$ $\pm$ 1.1}             &             \textbf{22.16 $\pm$ 1.1}             &      \textbf{26.86$^*$ $\pm$ 0.7}                     \\ \midrule
\textbf{ER$_{aug}$}  &                  81.30 $\pm$ 2.0       &   \textbf{77.71 $\pm$ 0.8}  &      \textbf{80.31 $\pm$ 0.9}                          &              47.97 $\pm$ 3.5    &     18.47 $\pm$ 2.0    &           31.75 $\pm$ 1.0                    \\ 
\textbf{ER$_{aug}$ + GMED}  &       \textbf{82.39$^*$ $\pm$ 3.7}       &      77.68 $\pm$ 0.8     &             80.30 $\pm$ 1.0                          &              \textbf{51.38$^*$ $\pm$ 2.2}        &  \textbf{18.63 $\pm$ 1.3}     &           \textbf{31.83 $\pm$ 0.8}     \\         

\bottomrule

\end{tabular}
}

\label{tab:fuzzy_full}
\vspace{-0.1cm}
\end{table*}

Table~\ref{tab:fuzzy_full} summarizes full results of experiments where we apply fuzzy boundaries between tasks. Experiments show that GMED generally improves performance on Split MNIST, Split CIFAR-10 and Split mini-ImageNet significantly when built upon ER, MIR, or ER$_{aug}$.

\section{Comparison to Task-Aware CL Methods}
\label{apdx:task_aware}
We additionally present the results of task-aware memory-based approaches in Table~\ref{tab:task_aware_tab}. We notice HAL was not as competitive as ER on Split MNIST, Split CIFAR-10, and Split mini-ImageNet that employs a class incremental learning setup --- while in the original work, the approach was mainly test in a domain-incremental learning setup~\citep{Ven2019ThreeSF}. GEM performs competitively on two of the datasets (Split MNIST and Rotated MNIST), where GEM+GMED could further slightly improve the performance. 
\begin{table*}[h]
\vspace{-0.1cm}
\caption{\textbf{Performance of Task-Aware Approaches.}
}
\label{tab:task_aware_tab}

\centering
\scalebox{0.66}{\begin{tabular}{@{}lcccccc@{}}
\toprule
\textbf{Methods / Datasets} & \textbf{\begin{tabular}[c]{@{}c@{}}Split\\ MNIST\end{tabular}}  & \textbf{\begin{tabular}[c]{@{}c@{}}Rotated\\ MNIST\end{tabular}} & \textbf{\begin{tabular}[c]{@{}c@{}}Split\\ CIFAR-10\end{tabular}} &  \textbf{\begin{tabular}[c]{@{}c@{}}Split\\ mini-ImageNet\end{tabular}} \\ \midrule
HAL &  77.92 $\pm$ 4.2 &  78.48 $\pm$ 1.5 & 32.06 $\pm$ 1.5  & 21.18 $\pm$ 2.1 \\
\midrule
GEM &  87.21 $\pm$ 1.3  & 78.40 $\pm$ 0.5  & 14.81 $\pm$ 0.4  &  5.92 $\pm$ 0.6 \\
GEM + GMED   &     87.69 $\pm$ 1.4                       &          78.62 $\pm$ 0.4         &      14.13 $\pm$ 0.3    &  5.81 $\pm$ 0.5  \\
\bottomrule
\end{tabular}
}

\end{table*}

\section{GMED without Writing Edits Back}
Table~\ref{tab:no_write_back} summarizes the result of discarding the editing performed after each time step: in these experiments, we replay the edited examples, but do not replace original examples in the memory with edited ones. The approach is noted as ER+GMED w/o writeback. We tune a separate set of editing stride and regularization strength for the method. The results indicate that ER+GMED w/o writeback achieves slightly lower performance on 5 out of 6 datasets (except Split CIFAR-10) compared to ER+GMED.
\begin{table*}[!htbp]
\centering
\caption{GMED without storing back edited examples, \ie, the algorithm replays edited examples, but does not update the original examples in the memory as edited ones. The results are shown after ER + GMED w/o writeback.}
\label{tab:no_write_back}
\scalebox{0.72}{
\begin{tabular}{@{}lcccccc@{}}
\toprule
\textbf{Methods / Datasets} & \textbf{\begin{tabular}[c]{@{}c@{}}Split\\ MNIST\end{tabular}} & \textbf{\begin{tabular}[c]{@{}c@{}}Permuted\\ MNIST\end{tabular}} & \textbf{\begin{tabular}[c]{@{}c@{}}Rotated\\ MNIST\end{tabular}} & \textbf{\begin{tabular}[c]{@{}c@{}}Split\\ CIFAR-10\end{tabular}} & \textbf{\begin{tabular}[c]{@{}c@{}}Split\\ CIFAR-100\end{tabular}} & \textbf{\begin{tabular}[c]{@{}c@{}}Split\\ mini-ImageNet\end{tabular}} \\ \midrule
\textbf{ER}  & 81.07 $\pm$ 2.5                                                & 78.85 $\pm$ 0.7                                                   & 76.71 $\pm$ 1.6                                                  & 33.30 $\pm$ 3.9                               &     20.41 $\pm$ 1.2                    & 25.92 $\pm$ 1.2                                                        \\
\textbf{ER + GMED}     & 82.67 $\pm$ 1.9                                           & 78.86 $\pm$ 0.7                                                  & 77.09 $\pm$ 1.3                                              & 34.84 $\pm$ 2.2                              &      20.93 $\pm$ 1.6          & 27.27 $\pm$ 1.8                                                    \\
\textbf{ER + GMED w/o writeback}    &  81.18 $\pm$ 2.6                                       & 78.49 $\pm$ 0.7                                                   &  76.88 $\pm$ 1.1                                 &   34.86 $\pm$ 2.7                              &     20.86 $\pm$ 1.6        & 27.20 $\pm$ 1.8                                                    \\
\bottomrule
\end{tabular}
}

\vspace{-0.1cm}
\end{table*}


\section{Applying the Full Dataset in Split MNIST}
In our main experiments, we sampled 1,000 training examples for Split MNIST following~\citep{Aljundi2018TaskFreeCL}. We further include the results of using the entire dataset for training in Table~\ref{tab:full_mnist}, and still see significant improvements.
\begin{table*}[h]

\caption{\textbf{Performance on Split MNIST constructed from the entire MNIST training set}. Note than in our main experiments, we sampled 1,000 examples per task. $^*$ and $^{**}$ indicates significant improvement with $p<0.1$ and $p<0.01$ respectively.
}
\label{tab:full_mnist}

\centering
\scalebox{0.7}{\begin{tabular}{@{}lcc@{}}
\toprule
\textbf{Methods} & \textbf{w/o GMED} & \textbf{w/ GMED}  \\
\midrule
ER & 86.61 $\pm$ 1.3 & \textbf{88.48$^{**}$ $\pm$ 1.0} \\
MIR & 89.18 $\pm$ 1.5 & \textbf{89.88$^{**}$ $\pm$ 1.1} \\
ER$_{aug}$ & 91.52 $\pm$ 1.5 & \textbf{92.30$^{*}$ $\pm$ 0.9} \\
\midrule
\end{tabular}
}
\end{table*}

\section{Tabular Results of Model Performance under Various Memory Sizes}
Figure~\ref{fig:mem_changes} in the main paper present model performance under various memory size setups with bar charts. In Table~\ref{tab:mem_change}, we further summarize results in tables, showing exact numbers of mean and standard deviation of performance.
We find the improvements on Split-MNIST and Split mini-ImageNet are significant-with <0.05 over all memory size setups. Improvements on Rotated MNIST and Split CIFAR-10 are also mostly significant. 
\begin{table*}[h]
\centering
\caption{\textbf{Performance of ER, GMED$+$ER, MIR, and GMED$+$MIR with various memory sizes}, shown in tables. $^*$ indicates significant improvement with $p<0.05$.}
\scalebox{0.60}{
\begin{tabular}{@{}lccccccccc@{}}
\toprule
\textbf{Dataset}       & \multicolumn{3}{c}{\textbf{Split MNIST}}            & \multicolumn{3}{c}{\textbf{Rotated MNIST}}          & \multicolumn{3}{c}{\textbf{Split CIFAR-10}}         \\ \cmidrule(r){1-1} \cmidrule(lr){2-4} \cmidrule(lr){5-7} \cmidrule(lr){8-10}
\textbf{Mem size}      & 100             & 200             & 500             & 100             & 200             & 500             & 100             & 200             & 500             \\ \midrule
\textbf{ER}       & 71.13 $\pm$ 1.5 & 77.85 $\pm$ 1.4 & 81.07 $\pm$ 2.5 & 63.35 $\pm$ 1.7 & 70.07 $\pm$ 1.2 & 76.71 $\pm$ 1.6 & 21.94 $\pm$ 1.5 & 26.03 $\pm$ 2.1 & 33.30 $\pm$ 3.9  \\
\textbf{ER+GMED}  & 73.90$^*$ $\pm$ 3.7  & 79.19$^*$ $\pm$ 1.6 & 82.67$^*$ $\pm$ 1.9 & 64.23$^*$ $\pm$ 1.5 & 70.71$^*$ $\pm$ 1.2 & 77.09 $\pm$ 1.3 & 22.89$^*$ $\pm$ 1.8 & 27.18$^*$ $\pm$ 2.0 & 34.84$^*$ $\pm$ 2.2 \\
\textbf{MIR}      & 73.53 $\pm$ 1.6 & 80.79 $\pm$ 2.5 & 85.72 $\pm$ 1.2 & 62.91 $\pm$ 1.1 & 69.95 $\pm$ 1.2 & 77.50 $\pm$ 1.6  & 23.09 $\pm$ 1.2 & 28.78 $\pm$ 2.0 & 34.42 $\pm$ 2.4 \\
\textbf{MIR+GMED} & 77.09$^*$ $\pm$ 2.5 & 83.52$^*$ $\pm$ 1.1 & 86.52$^*$ $\pm$ 1.4 & 64.39$^*$ $\pm$ 0.4 & 70.62$^*$ $\pm$ 1.2 & 79.08$^*$ $\pm$ 0.8 & 25.87$^*$ $\pm$ 1.9 & 28.89 $\pm$ 1.8 & 36.17 $\pm$ 2.5 \\
\midrule
\textbf{Dataset}       & \multicolumn{3}{c}{\textbf{Split CIFAR-100}}        & \multicolumn{3}{c}{\textbf{Split mini-ImageNet}}    & \multicolumn{3}{c}{\multirow{6}{*}{\cellcolor{na}}}               \\ \cmidrule(r){1-1} \cmidrule(lr){2-4} \cmidrule(lr){5-7} \cmidrule(lr){8-10}
\textbf{Mem size}      & 2000            & 5000            & 10000           & 5000            & 10000           & 20000           &  \multicolumn{3}{c}{\cellcolor{na}} \\ \midrule
\textbf{ER}       & 12.31 $\pm$ 1.0 & 16.86 $\pm$ 0.4 & 20.41 $\pm$ 1.2 & 18.92 $\pm$ 0.9 & 25.92 $\pm$ 1.2 & 29.93 $\pm$ 1.9 & \multicolumn{3}{c}{\cellcolor{na}}                                \\                            
\textbf{ER+GMED}  & 12.65 $\pm$ 0.8 & 17.40$^*$ $\pm$ 0.7  & 20.93 $\pm$ 1.6 & 21.36$^*$ $\pm$ 0.4 & 27.27$^*$ $\pm$ 1.8 & 30.60$^*$ $\pm$ 1.8  & \multicolumn{3}{c}{\cellcolor{na}}                                \\
\textbf{MIR}      & 11.85 $\pm$ 0.8 & 18.36 $\pm$ 1.5 & 20.02 $\pm$ 1.7 & 17.47 $\pm$ 1.0 & 25.21 $\pm$ 2.2 & 30.08 $\pm$ 1.2 & \multicolumn{3}{c}{\cellcolor{na}}                                \\ 
\textbf{MIR+GMED} & 12.32 $\pm$ 0.8 & 18.60 $\pm$ 1.1  & 21.22$^*$ $\pm$ 1.0 & 20.31$^*$ $\pm$ 0.8 & 26.50$^*$ $\pm$ 1.3  & 31.33$^*$ $\pm$ 1.6 & \multicolumn{3}{c}{\cellcolor{na}}                                \\ \bottomrule
\end{tabular}
}
\label{tab:mem_change}
\end{table*}

\begin{table*}[]
\centering
\caption{Performance of optimal edits in Sec.~\ref{ssec:gmed_abl_optimal} and their comparison to ER and GMED.}
\label{tab:optimal_performance}
\scalebox{0.72}{
\begin{tabular}{@{}lcccc@{}}
\toprule
\textbf{Methods/Datasets} & \textbf{Split MNIST} & \textbf{Rotated MNIST} & \textbf{Split CIFAR-10} & \textbf{Split mini-ImageNet} \\ \midrule
\textbf{ER}                        & 80.14$\pm$3.2        & 76.71$\pm$1.6          & 33.30$\pm$3.9     &  25.92$\pm$1.2        \\
\textbf{ER+GMED}                   & 82.67$\pm$1.9        & 77.09$\pm$1.3          & 34.84$\pm$2.2      &  27.27$\pm$1.8    \\
\textbf{ER+Optimal Editing}                & 83.40$\pm$2.6        & 77.73$\pm$1.3          & 35.04$\pm$2.6      & 27.01$\pm$1.6    \\ \bottomrule
\end{tabular}}
\end{table*}

\section{Performance of Optimal Editing}
\label{apdx:optimal_editing}
Table~\ref{tab:optimal_performance} summarizes the performance of optimal editing (ER+Optimal) discussed in Sec.~\ref{ssec:gmed_abl_optimal} compared to ER and ER+GMED. We notice that ER+Optimal outperforms ER+GMED on Split MNIST and Rotated MNIST, slightly improves on Split CIFAR-10, but does not improve on Split mini-ImageNet. The hyperparameters are not tuned extensively on mini-ImageNet because optimal editing is very expensive to compute, requiring to compute forgetting over a large sample of early examples every training step, even on moderately-sized datasets.

We hereby note that our ER+Optimal does not necessarily achieve an upper-bound performance of memory editing, as it computes optimal edit for only the coming \textit{one} training step over the data stream. Because the edited examples may in turn affect training dynamics in future training steps, it is hard to derive an exact upper bound performance of memory editing.

\begin{table*}[]
\caption{Results when editing an additional number of examples per step in ER+GMED.}
\label{tab:edit_extra_examples}
\centering
\scalebox{0.72}{
\begin{tabular}{@{}lcccc@{}}
\toprule
\textbf{Methods/Dataset}       & \textbf{Split MNIST}   & \textbf{Rotated MNIST} & \textbf{Split CIFAR-10} & \textbf{Split mini-ImageNet} \\ \midrule
\textbf{ER+GMED (10 examples)} & 82.67$\pm$1.9 & 77.09$\pm$1.3 & 34.84$\pm$2.2  & 27.27$\pm$1.8       \\
\textbf{+ 3 examples}          & 83.00$\pm$2.3 & 77.01$\pm$1.6 & 34.87$\pm$2.9  & 27.36$\pm$1.9       \\
\textbf{+ 5 examples}          & 82.78$\pm$2.3 & 76.98$\pm$1.6 & 35.31$\pm$2.2  & 28.00$\pm$2.0       \\
\textbf{+ 10 examples}         & 82.21$\pm$2.2 & 76.51$\pm$1.4 & 34.86$\pm$2.8  & 27.17$\pm$1.8       \\
\textbf{+ 50 examples}         & 81.42$\pm$2.6 & 76.50$\pm$1.3 & 31.79$\pm$2.6  & 27.10$\pm$1.9   \\   \bottomrule
\end{tabular}
}
\end{table*}

\section{Editing Extra Examples}
Table~\ref{tab:edit_extra_examples} summarizes the results when we edit an additional number of examples per task. We randomly sample additional examples from the memory, perform edits, and writeback, without replaying them. We notice that, over all datasets, replaying a small number of extra examples improve the performance. However, the performance drops when too many examples are edited. We hypothesize replaying additional examples per step has a similar effect to increasing the stride of edits performed per example. As such, when required, the number of additional examples to edit can be an additional tunable hyperparameter in our approach.

\section{Prediction Change Rate of Edited Examples}

\begin{table*}[]
\caption{Prediction change rate of edited examples compared to the corresponding original examples.}
\centering
\label{tab:prediction_change_rate}
\scalebox{0.7}{
\begin{tabular}{@{}lcccc@{}}
\toprule
Dataset       & Split MNIST   & Rotated MNIST & Split CIFAR-10 & Split mini-ImageNet \\ \midrule
Prediction Change Rate & 10.2\% & 2.4\% & 5.1\%  & 5.5\%       \\ \bottomrule
\end{tabular}}
\end{table*}

We show the percentage of changed predictions after example edits in Table~\ref{tab:prediction_change_rate}. Specifically, when the training completes, we classify examples stored in the memory (which have experienced editing) and the corresponding original examples with the model. We then compute ``prediction change rate'' as the portion of changed predictions over all examples stored in the memory. The prediction change implies the edited examples are more adversarial, or they can be simply artifacts to the model. However, we notice that such prediction change rate is positively correlated with performance improvement of GMED over four datasets. It implies such adversarial or artifact examples are still beneficial to reducing forgetting.

\section{Building Upon Other Memory Population Strategies}

\begin{table*}[]
\caption{Prediction change rate of edited examples compared to the corresponding original examples.}
\centering
\label{tab:gss_gmed}
\scalebox{0.7}{
\begin{tabular}{@{}lcc@{}}
\toprule
\textbf{Method/Dataset}       & \textbf{Split MNIST}   & \textbf{Rotated MNIST}  \\ \midrule
\textbf{ER+GSS-Greedy} & 83.70$\pm$1.0 & 73.80$\pm$1.6       \\ 
\textbf{ER+GSS-Greedy+GMED}  & 84.62$\pm$1.2 & 74.47$\pm$1.3  \\  \midrule
$p$-value & 0.016 & 0.18 \\ \bottomrule
\end{tabular}}
\end{table*}

In all our main experiments, we used reservoir sampling to populate the replay memory. We further experiment with the greedy variant of Gradient-based Sample Selection strategy (GSS-Greedy)~\citep{Aljundi2019GradientBS} on feasibly-sized datasets, which tries to populate the memory with more diverse examples. We report the results in Table~\ref{tab:gss_gmed}. The results show GMED could still bring modest improvements when built upon GSS-Greedy.

\section{Ethics Statement: Societal Impact}
\label{sec:supp_ethics_statement}
In this work, we extend existing memory-based continual learning methods to allow gradient-based editing. We briefly discuss some of the societal impacts in this section.

\textbf{Societal Impact (Positive):} 
Improvement in Continual learning, and  in particular, the online version of continual learning, can lead to potential benefits in wide variety of machine-learning tasks dealing with streaming data. With novel information being introduced by the day, it is imperative that models are not re-trained on the entire data, instead adapt and take into account the streaming data without forgoing previously learned knowledge. As an example, BERT trained on 2018 language data would be less useful for current events, but still be useful for commonsense knowledge information and as such it would be extremely beneficial if one could train over streaming current events while retaining other knowledge.

Towards this goal, a primary advantage of GMED is its integration with other memory-based continual learning frameworks. As a result, we expect advances in memory-based methods to be complemented by GMED, incurring only a minor computational overhead -- a key requirement for deploying any online continual learning algorithm in the wild.

\textbf{Societal Impact (Negative):} Caution must be taken in deploying our continual learning algorithms in the wild. This is because, CL algorithms at present, are validated solely on small datasets with synthetically curated streams. 
In the wild, the examples in the continual stream can have unexpected correlations which are not apparent in image-classification only streams.

Another key issue with continual learning is that once a knowledge is learned it is difficult to know whether it has been completely unlearned or still retained in the neural network which can be probed later. 
This could happen for say hospital records where patient confidentiality is needed, but were used by the continual learning model. Deleting such records is non-trivial for usual machine learning models but have dire consequences in the continual learning domain where the original stream can no longer be accessed. 


\end{document}